\documentclass[journal]{IEEEtran}
\usepackage{amsmath,amsfonts}
\usepackage{amssymb}
\usepackage{dsfont}
\usepackage{algorithmic}
\usepackage{algorithm}
\usepackage{array}
\usepackage{textcomp}
\usepackage{stfloats}
\usepackage{url}
\usepackage{verbatim}
\usepackage{graphicx}
\usepackage{bm}
\usepackage{newfloat}
\usepackage{listings}
\usepackage{booktabs}
\usepackage{multirow}
\usepackage{mathrsfs}
\usepackage{amssymb}
\usepackage{amsthm}
\usepackage{dsfont}
\usepackage{xcolor}
\usepackage{cite}

\usepackage{adjustbox}
\usepackage[caption=false,font=normalsize,labelfont=rm,textfont=rm]{subfig}
\usepackage{amsfonts}
\usepackage{epstopdf}

\usepackage{amssymb}
\usepackage{cases}
\usepackage{color}
\usepackage{ulem}
\usepackage{diagbox}
\usepackage{colortbl}
\definecolor{mygray}{gray}{.9}
\usepackage{stfloats}
\usepackage{pifont}

\usepackage{hyperref}
\hypersetup{
    colorlinks=true,
    linkcolor=blue,
    filecolor=magenta,      
    urlcolor=cyan,	
    pdfpagemode=FullScreen,
    citecolor=blue,
    }
\newtheorem{proposition}{Proposition}
\newtheorem{theorem}{Theorem}
\newtheorem{assumption}{Assumption}
\usepackage{booktabs}
\begin{document}
\newcommand{\removelatexerror}{\let\@latex@error\@gobble}
\makeatother
\title{Spatial-Aware Conformal Prediction for Trustworthy Hyperspectral Image Classification}
\author{Kangdao Liu,
        Tianhao Sun, 
        Hao Zeng, Yongshan Zhang,~\IEEEmembership{Member, IEEE},
         Chi-Man Pun,~\IEEEmembership{Senior Member, IEEE},
        and Chi-Man Vong,~\IEEEmembership{Senior Member, IEEE}
        \thanks{This work has been submitted to the IEEE for possible publication. Copyright may be transferred without notice, after which this version may no longer be accessible.}
        \thanks{Kangdao liu and Tianhao Sun contributed equally to this work.}
        \thanks{Corresponding authors: Chi-Man Vong; Chi-Man Pun.}
        \thanks{Kangdao Liu is with the Department of Computer and Information Science, University of Macau, Macau 999078, China, and also with the Department of Statistics and Data Science, Southern University of Science and Technology, Shenzhen 518055, China (e-mail: kangdaoliu@gmail.com).}
        \thanks{Tianhao Sun, Chi-Man Pun and Chi-Man Vong are with the Department of Computer and Information Science, University of Macau, Macau 999078, China (e-mail: mc25124@um.edu.mo;  cmpun@um.edu.mo; cmvong@um.edu.mo).}
        \thanks{Hao Zeng is with the Department of Statistics and Data Science, Southern University of Science and Technology, Shenzhen 518055, China (e-mail: zengh@sustech.edu.cn).}
                \thanks{Yongshan Zhang is with the School of Computer Science, China University of Geosciences, Wuhan 430074, China(e-mail: yszhang@cug.edu.cn).}
}
\maketitle
\begin{abstract}

Hyperspectral image (HSI) classification involves assigning unique labels to each pixel to identify various land cover categories.
While deep classifiers have achieved high predictive accuracy in this field, they lack the ability to rigorously quantify confidence in their predictions.
 Quantifying the certainty of model predictions is crucial for the safe usage of predictive models, and this limitation restricts their application in critical contexts where the cost of prediction errors is significant.
To support the safe deployment of HSI classifiers, we first provide a theoretical proof establishing the validity of the emerging uncertainty quantification technique, conformal prediction, in the context of HSI classification.
We then propose a conformal procedure that equips any trained HSI classifier with trustworthy prediction sets, ensuring that these sets include the true labels with a user-specified probability (e.g., 95\%).
Building on this foundation, we introduce Spatial-Aware Conformal Prediction (\texttt{SACP}), a conformal prediction framework specifically designed for HSI data. 
This method integrates essential spatial information inherent in HSIs by aggregating the non-conformity scores of pixels with high spatial correlation, which effectively enhances the efficiency of prediction sets.
Both theoretical and empirical results validate the effectiveness of our proposed approach.
The source code is available at \url{https://github.com/J4ckLiu/SACP}.

\end{abstract}
\begin{IEEEkeywords}
Hyperspectral image classification, uncertainty quantification, conformal prediction
\end{IEEEkeywords}

\section{Introduction}
Hyperspectral imaging technology captures spectral information across a wide wavelength range, providing more detailed pixel features compared to traditional three-channel color images \cite{zhang:TIP24, zhang:TMM24}.
Hyperspectral image (HSI) classification, a fundamental task in this domain, involves assigning unique labels to each pixel to identify various land cover categories.
While deep classifiers have demonstrated high accuracy in this field \cite{dong2022weighted, zhao2023hyperspectral, sun2021supervised, zheng2022rotation, xi2022few}, performance metrics like accuracy alone are insufficient for their deployment in safety-critical contexts.
In these scenarios, it is also crucial to quantify the certainty of predictions.
This need is amplified by the significant applications of HSIs, including surveillance, threat detection \cite{kutuk2023ground}, mineral detection \cite{murphy2013consistency}, agriculture \cite{zhang2016crop}, environmental protection \cite{bioucas2013hyperspectral}, and defense \cite{coffey2015hyperspectral}, where prediction errors could have serious consequences.
For example, in threat detection or mineral exploration, incorrect predictions could result in misidentifying potential threats or valuable resources, leading to severe repercussions. 
This emphasizes the necessity of advancing methodologies that rigorously quantify the uncertainty of HSI classifiers to ensure their safe deployment.

Various techniques have been developed to estimate and incorporate uncertainty into predictive models, including confidence calibration \cite{guo2017calibration}, Monte Carlo Dropout \cite{gal2016dropout}, and Bayesian neural networks \cite{smith2013uncertainty}. 
Additionally, there are uncertainty quantification strategies specifically tailored to HSI classification.  
For instance, PL-CNN \cite{he2022toward} reduces both predictive and label uncertainty for more reliable predictions, achieving high classification accuracy while lowering the Expected Calibration Error (ECE) \cite{guo2017calibration}.
Another approach leverages Bayesian deep learning to estimate prediction uncertainty while enhancing the structural characteristics of HSI classifiers \cite{he2023bayesian}. 
Furthermore, uncertainty quantification techniques specifically applied to graph-based HSI classifiers have been explored in \cite{yu2023uncertainty}. This study adapts advanced uncertainty quantification models, initially developed for node classification in graphs, to the context of HSI classification. Collectively, these techniques contribute valuable insights into quantifying or reducing decision uncertainty. However, they lack theoretical guarantees of model performance \cite{huang2023conformal}, underscoring the importance of conformal prediction \cite{vovk2005algorithmic, balasubramanian2014conformal, angelopoulos2021gentle}.

Unlike the techniques discussed earlier, conformal prediction—an emerging uncertainty quantification methodology—typically operates as post-hoc processing for the outputs of trained classifiers. This statistical framework can be applied to any predictive model, transforming the output probabilities into prediction sets while ensuring that the true label is included with a user-specified coverage probability. Notably, the validity of the prediction sets is assured in a distribution-free manner, providing explicit, non-asymptotic guarantees without reliance on any distributional or model assumptions \cite{vovk2005algorithmic}. Formally, this guarantee can be expressed as follows:
\begin{equation}
\label{eq:target}
    \mathbb{P}(Y_{\text{test}} \in \mathcal{C}(X_{\text{test}})) \geq 1 - \alpha,
\end{equation}
where \( Y_{\text{test}} \) is the true label, \( \mathcal{C}(X_{\text{test}}) \) is the generated prediction set, and \( \alpha \) is the user-specified error rate. 
The prediction set will consistently include the true labels of the inputs with the specified error frequency.
The strong theoretical foundation of conformal prediction supports its efficacy across various applications, including classification \cite{sadinle2019least, huang2023conformal}, regression \cite{romano2019conformalized}, and specialized areas such as image generative models \cite{horwitz2022conffusion}, robotic control \cite{wang2023conformal}, graph neural networks \cite{zargarbashi2023conformal, huang2024uncertainty}, and large language models (LLMs) \cite{cherian2024large, su2024api, quach2023conformal}.

Despite its success in numerous domains, conformal prediction has remained unexplored in the context of HSI data.
There are two primary challenges that need to be addressed for the valid and effective implementation of conformal prediction in HSI classification.
Firstly, unlike traditional classification tasks, HSI classifiers typically require access to the entire HSI, including test instances, during training. This practice could potentially undermine the validity of conformal prediction, which relies on the exchangeability of calibration and test data \cite{vovk2005algorithmic}. 
Therefore, it is imperative to theoretically examine whether the validity of conformal prediction is maintained in the context of HSI classification.  Secondly, the performance of conformal prediction algorithms is dictated by the non-conformity score function once the base classifier is determined.  Although many effective non-conformity score functions have been proposed—such as Adaptive Prediction Sets (APS) \cite{romano2020classification}, which calculates the non-conformity scores by accumulating descending softmax values, and Regularized Adaptive Prediction Sets (RAPS) \cite{angelopoulos2020uncertainty}, which enhances the efficiency of conformal prediction sets through regularization based on label ranks—these non-conformity score functions generally neglect the critical spatial information inherent in HSIs. 
This oversight can lead to suboptimal prediction sets with unnecessarily large sizes. 
In summary, addressing the following two questions is essential for the effective application of conformal prediction in HSI classification:
\begin{itemize}
\item \textbf{Does the validity of conformal prediction persist in the context of HSI classification?}
\item \textbf{How can spatial information be utilized to enhance the performance of conformal prediction?}
\end{itemize}

In this paper, we first present a theoretical analysis that confirms the validity of conformal prediction in HSI classification, even when classifiers are exposed to pixels from both calibration and test sets during training. 
Building on this foundation, we propose a framework for integrating conformal prediction sets into HSI classifiers, ensuring the inclusion of the true label with a user-specified coverage probability. 
Furthermore, we theoretically illustrate that incorporating neighborhood information can enhance the statistical efficiency of prediction sets.
Motivated by this insight, we introduce \textbf{Spatial-Aware Conformal Prediction} (\texttt{SACP}), which refines standard conformal prediction by aggregating the non-conformity scores of pixels exhibiting high spatial correlation. 
This approach effectively leverages the spatial information inherent in HSIs to improve the performance of conformal prediction. Moreover, we present a theoretical demonstration of the validity of \texttt{SACP}, confirming that it guarantees coverage as defined in Eq. \eqref{eq:target}.

We conduct extensive evaluations on standard HSI classification benchmarks, including Indian Pines, Pavia University, and Salinas. 
The experimental results confirm the validity of our proposed conformal framework (standard conformal prediction) and \texttt{SACP} for HSI classification: the generated prediction sets consistently include the true label of data with a user-specified coverage probability. 
Furthermore, \texttt{SACP} significantly enhances the efficiency of various non-conformity score functions while maintaining satisfactory coverage rates across different scenarios, outperforming standard conformal prediction in this context. 
This finding underscores the critical importance of incorporating spatial information into conformal prediction. 
Notably, \texttt{SACP} demonstrates its effectiveness with both simple classifiers, such as CNN-based models, and advanced classifiers, including Transformer-based models, thereby highlighting the adaptability of our proposed approach.

Our main contributions are highlighted as follows:
\begin{itemize}
    \item To the best of our knowledge, we are the first to explore the application of the emerging uncertainty quantification technique, conformal prediction, to HSI data. We present a thorough theoretical proof that establishes the validity of conformal prediction for HSI classification, thereby opening a new avenue for the safe utilization of HSI.
    \item We introduce a framework that equips HSI classifiers with reliable conformal prediction sets, ensuring the inclusion of true labels with a user-specified probability.
    \item We provide a theoretical demonstration that incorporating spatial information can enhance the statistical efficiency of prediction sets. Building on this insight, we propose \texttt{SACP}, a framework that leverages the inherent spatial information in HSIs to improve the performance of conformal prediction. We theoretically demonstrate the validity of \texttt{SACP} by confirming its coverage guarantee.
    \item Extensive experiments validate the effectiveness of our proposed method. The emprical results confirm the validity of conformal prediction for HSI classification. Moreover, \texttt{SACP} outperforms standard conformal prediction by reducing the average prediction set size while maintaining satisfactory coverage rates across different scenarios.
\end{itemize}

\section{Background}
\subsection{Conformal Prediction}
 \paragraph{Notation} In this section, we consider the multi-class classification task involving \( K \) classes. Let \(\mathcal{X} \subset \mathbb{R}^d\) represent the input space, and \(\mathcal{Y} = \{1, \ldots, K\}\) denote the label space. Let \((\mathit{X}, \mathit{Y})\) be a random data pair drawn from the joint distribution \(\mathcal{P_{XY}}\). We use \( f_{\theta} : \mathcal{X} \rightarrow \mathbb{R}^K \) to denote a pre-trained classifier with parameters \(\theta\). For the input \(\bm{x}\), \( f_{\theta, y}(\bm{x}) \) represents the \( y \)-th element of the logits vector \( f_{\theta}(\bm{x}) \). The conditional probability of class \( y \) is approximated by:
\begin{equation*}
    \hat{\pi}_{\theta,y}(\bm{x})=\sigma(f_\theta(\bm{x}))_y=\frac{e^{f_{\theta,y}(\bm{x})}}{\sum_{i=1}^{\mathit{K}} e^{f_{\theta,i}(\bm{x})}},
\end{equation*}
where \(\sigma\) denotes the softmax function and \(\hat{\pi}_{\theta}(\bm{x}) = (\hat{\pi}_{\theta,1}(\bm{x}), \hat{\pi}_{\theta,2}(\bm{x}), \dots, \hat{\pi}_{\theta,K}(\bm{x}))\) represents the softmax probability distribution. Typically, the classification result is obtained by \(\hat{y} = \mathrm{argmax}_{y \in \mathcal{Y}} \hat{\pi}_{\theta,y}(\bm{x})\).

\paragraph{Pipeline} Instead of predicting a single label based on the output probabilities of a classifier, conformal prediction constructs a set-valued mapping \(\mathcal{C} : \mathcal{X} \rightarrow 2^{\mathcal{Y}}\) that satisfies the following marginal coverage property:
\begin{equation*}
    \mathbb{P}(\mathit{Y} \in \mathcal{C}(\mathit{X})) \geq 1 - \alpha,
\end{equation*}
where $\alpha\in (0,1)$ denotes the user-specified error rate and $\mathcal{C}(\mathit{X})$ is the prediction set, which is a subset of $\mathcal{Y}$.

Before deployment, conformal prediction starts with a calibration step using a separate calibration set, \(\mathcal{D}_{\text{cal}} := \{(\boldsymbol{x}_i, y_i)\}_{i = 1}^{n}\), drawn independently from the distribution \(\mathcal{P}_{\mathcal{XY}}\). Specifically, we compute the non-conformity score \(s_i = \mathit{S}(\boldsymbol{x}_i, y_i)\) for each sample in \(\mathcal{D}_{\text{cal}}\), which measures how $y_i$ ``conforms" to the prediction at $\boldsymbol{x}_i$. The $1-\alpha$ quantile of the scores $\{s_i\}_{i = 1}^{n}$ is then determined as the threshold $\tau$. Formally, \(\tau\) is obtained as follows:
\begin{equation}
\label{eq:calculate_tau}
    \tau = \inf \left\{ s \mid \frac{\left|\{i  : s_i \leq s\}\right|}{n} 
    \geq \frac{\lceil (n + 1)(1 - \alpha) \rceil}{n} \right\}.
\end{equation}
Once the threshold \(\tau\) is determined, the prediction set for the test instance \(\bm{x}_{n+1}\) at the error rate \(\alpha\) is generated by:
\begin{equation}\label{equ4}
    \mathcal{C}_{1-\alpha}(\bm{x}_{n+1};\tau):=\{y\in\mathcal{Y} \mid \mathit{S}(\bm{x}_{n+1},y) \leq \tau\},
\end{equation}
where \(\mathcal{C}_{1-\alpha}(\bm{x}_{n+1};\tau)\) includes all labels for which their non-conformity score \(\mathit{S}(\bm{x}_{n+1}, y)\) does not exceed the threshold \(\tau\).
As stated in the following theorem, the prediction sets generated by Eqs. \eqref{eq:calculate_tau} and \eqref{equ4} are guaranteed to achieve finite-sample marginal coverage of the true labels.
\begin{theorem}{\cite{vovk2005algorithmic}}\label{thm:cp}
Let \((\bm{x}_i, y_i)_{i=1}^{n}\) and \((\boldsymbol{x}_{n+1}, y_{n+1})\) be exchangeable data samples. Suppose \(\tau\) is calculated by Eq. \eqref{eq:calculate_tau} using \((\bm{x}_i, y_i)_{i=1}^{n}\), and the prediction set for \((\boldsymbol{x}_{n+1}, y_{n+1})\) is given by Eq. \eqref{equ4}. Then, the coverage guarantee is as follows:
    \begin{equation*}
        \mathbb{P}\left(y_{n+1} \in \mathcal{C}_{1-\alpha}\left(\boldsymbol{x}_{n+1}; \tau\right)\right) \geq 1 - \alpha.
    \end{equation*}
\end{theorem}
Notably, the validity is ensured in a distribution-free manner, offering explicit, non-asymptotic guarantees without relying on distributional or model assumptions. Consequently, as long as the exchangeability assumption holds, the generated prediction sets will have a coverage guarantee. This highlights the flexibility of conformal prediction, as it can be applied to any \textbf{pre-trained classifier} when the assumption is satisfied.
\paragraph{Non-conformity score function} Once the base classifier is determined, the effectiveness of the resulting prediction set depends critically on the chosen non-conformity score function $S$. Two popular options are Adaptive Prediction Sets (APS) and Regularized Adaptive Prediction Sets (RAPS). APS calculates the non-conformity scores by accumulating the softmax probabilities, sorted in descending order:
\begin{equation*}
    S_{\text{APS}}(\bm{x},y)\!=\!\sum_{y_i \ne y} \hat{\pi}_{\theta,y_i} (\boldsymbol{x}) \cdot \mathds{1}_{\{\hat{\pi}_{\theta,y_i} (\boldsymbol{x}) > \hat{\pi}_{\theta,y}(\boldsymbol{x})\}}  + u\cdot\hat{\pi}_{\theta,y}(\boldsymbol{x}),
\end{equation*}
where \(u\) is an independent random variable following a uniform distribution on \([0, 1]\). However, the softmax probabilities typically exhibit a long-tailed distribution, which facilitates the inclusion of tail classes in the prediction sets. 
To address this issue, RAPS \cite{angelopoulos2020uncertainty} refines the non-conformity scores by incorporating regularization for the label rankings. Through this regularization, RAPS ensures that labels with higher rankings are associated with larger scores, defined as follows:
\begin{equation*}
\begin{array}{r}
         S_{\textrm{RAPS}}(\boldsymbol{x}, y) =    S_{\textrm{APS}}(\boldsymbol{x}, y) + \lambda\cdot(o(y,\hat{\pi}_\theta(\boldsymbol{x})) - k)^{+},
\end{array}
\end{equation*}
where \(o(y,\hat{\pi}_\theta(\boldsymbol{x}))\) is the label ranking of \(y\), \(\lambda\) and \(k\) are hyperparameters, and \((z)^{+}\) denotes the positive part of \(z\). Yet, the non-conformity score of RAPS still involves unreliable softmax probabilities, leading to suboptimal performance in conformal prediction. To solve it, Sorted Adaptive Prediction Sets (SAPS) \cite{huang2023conformal} reduces the emphasis on probabilities by retaining only the maximum softmax probability and discarding the others. Formally, SAPS is defined as follows:
\begin{equation*}
    S_{\text{SAPS}}(\boldsymbol{x}, y) = \left\{
    \begin{array}{l}
    u \cdot \hat{\pi}_{\theta, \max }(\boldsymbol{x}), \quad \text{if } o(y, \hat{\pi}_{\theta}(\boldsymbol{x})) = 1, \\
    \hat{\pi}_{\theta, \max }(\boldsymbol{x}) + (o(y, \hat{\pi}_{\theta}(\boldsymbol{x})) - 2 + u) \cdot \lambda,\text{else},
    \end{array}
    \right.
\end{equation*}
where $\lambda$ balances the weight of ranking information, $\hat{\pi}_{\theta, \max }(\boldsymbol{x})$ denotes the maximum softmax probability of instance $\bm{x}$, and $u$ is a random variable uniformly distributed on $[0, 1]$. \textbf{It can be concluded that these score functions depend exclusively on softmax probabilities and label ranks.}

\paragraph{Evaluation Metric}\label{back:cp}We evaluate the generated prediction sets using three key metrics: marginal coverage rate (Coverage), average size of the prediction sets (Size), and Size-Stratified Coverage Violation (SSCV). Coverage measures the percentage of samples whose sets contain true labels:
\begin{equation*}
    \text{Coverage} = \frac{1}{|\mathcal{D}_{\text{test}}|} \sum_{(\boldsymbol{x},y) \in \mathcal{D}_{\text{test}}} \mathds{1}_{\{y \in \mathcal{C}(\bm{x})\}},
\end{equation*}
which should ideally match the pre-defined coverage rate \(1 - \alpha\). Size measures the average length of prediction sets. When coverage is satisfied, a smaller size indicates higher statistical efficiency and is more desirable. Formally, Size is defined as:
\begin{equation*}
    \text{Size} = \frac{1}{|\mathcal{D}_{\text{test}}|} \sum_{(\boldsymbol{x},y) \in \mathcal{D}_{\text{test}}} |\mathcal{C}(\boldsymbol{x})|.
\end{equation*}
SSCV \cite{angelopoulos2020uncertainty} reflects the conditional coverage rate and is formally defined as:
\begin{equation*}
    \text{SSCV} = 100 \times \sup_{j} \left| (1 - \alpha) - \frac{\left|\{i : y_i \in \mathcal{C}(\boldsymbol{x}_i) \text{ and } i \in \mathcal{J}_j\}\right|}{|\mathcal{J}_j|} \right|,
\end{equation*}
where \(\mathcal{J}_j\) denotes the partitioned sets, with prediction sets categorized according to their sizes. This metric quantifies the maximum deviation between the observed coverage rate and \(1 - \alpha\) across the various size categories.

\subsection{HSI Classification} Let \(\mathcal{I} \in \mathbb{R}^{H \times W \times U}\) be the HSI to classify, where \(H\) and \(W\) denote the height and width in pixels, and \(U\) denotes the number of spectral bands. Let \(M = HW\) be the total number of pixels. Typically, HSI classification aims to predict the label \( y \in \mathcal{Y} \) for each pixel in the image, given that a random subset of labels known. The pixels with unknown labels appear randomly at any location in the HSI. When employing a deep classifier, the \((i, j)\)-th pixel is generally considered as the center, with information from the small cuboid \(B \in \mathbb{R}^{P \times P \times U}\) utilized to extract the joint spectral-spatial feature representation for the central pixel \cite{imani2020overview}\footnote{In prior research, the terms ``cuboids'', ``cubes'', and ``patches'' are often used interchangeably to describe local regions of HSIs. For clarity, this work consistently uses the term ``cuboids''.}. Here, \(P\) denotes the patch size of $B$. Correspondingly, the dataset constructed from \(\mathcal{I}\) comprises \(M\)  cuboids, defined as follows:
\begin{equation*}
    \mathcal{D}_{\mathcal{I}} := \{(B_{i}, y_{i}) \mid 1 \leq i \leq M\},
\end{equation*}
where \( y_{i} \) represents the label of the central pixel of \( B_{i} \). Here, $\{(B_{i}, y_{i})\}_{i=1}^{M}$ are exchangeable to $\mathcal{D}_{\mathcal{I}}$. Typically, the HSI dataset is partitioned into \(\mathcal{D}_{\text{train}}\), \(\mathcal{D}_{\text{val}}\), and \(\mathcal{D}_{\text{test}}\) for training, validation, and evaluation of the deep classifier. Essentially, the HSI classifier is engineered to accurately predict the label of the central pixel within each cuboid in \(\mathcal{D}_{\text{test}}\).

\section{Validity of Conformal Prediction on HSI}
\subsection{Theoretical Analysis}
As outlined in Theorem \ref{thm:cp}, conformal prediction requires that calibration and test samples be exchangeable. However, HSI classifiers typically operate on pixel cuboids as the fundamental unit of training, rather than individual pixels, and thus require access to the entire HSI during training.  Therefore, when partitioning \(\mathcal{D}_{\mathcal{I}}\) into \(\mathcal{D}_{\text{train}}\), \(\mathcal{D}_{\text{cal}}\), and \(\mathcal{D}_{\text{test}}\), cuboids in \(\mathcal{D}_{\text{train}}\) may overlap with those in \(\mathcal{D}_{\text{cal}}\) or \(\mathcal{D}_{\text{test}}\) due to shared pixels, which potential breaks the validity of conformal prediction. In this section, we demonstrate that, under a general permutation-invariance condition, the exchangeability of the non-conformity scores remains valid even when pixels from both the calibration and test sets are used during training. This theoretical analysis conclusively establishes the validity of applying conformal prediction to HSI classification.
\begin{assumption}
\label{ass:exchange}
For any instance cuboid \( B \) and any label \( y \in \mathcal{Y} \), the non-conformity score \( S(B, y) \) is invariant under any permutation of the union \( \mathcal{D}_{\text{cal}} \cup \mathcal{D}_{\text{test}} \).
\end{assumption}
Assumption \ref{ass:exchange} implies that the non-conformity scores for any instance-label pairs should remain unaffected by the choice of \( \mathcal{D}_{\text{cal}} \), thereby imposing a permutation-invariance condition on the training of the HSI classifier. This ensures that the output probabilities and, consequently, the non-conformity scores are invariant to permutations of the calibration and test data of the HSI. \textbf{HSI classifiers typically satisfy this assumption, as they do not take into account the ordering of pixels (or cuboids) during training.} Based on this fact, we then demonstrate that any  HSI classifier can produce valid prediction sets and offer a test-time coverage guarantee.
\begin{theorem}
\label{theo:exchange}
Let \(\{(B_i, y_i)\}_{i=1}^n\) and \(\{(B_{n+j}, y_{n+j})\}_{j=1}^m\) denote the calibration and test sets, respectively. Assume that the non-conformity score function \(S\), derived from the pre-trained HSI classifier, satisfies Assumption \ref{ass:exchange}. Define the non-conformity scores for the calibration and test sets as \(\{s_i\}_{i=1}^n\) and \(\{s_{n+j}\}_{j=1}^m\), where \(s_i := S(B_i, y_i)\). Under these conditions, the scores \(\{s_i\}_{i=1}^n\) and \(\{s_{n+j}\}_{j=1}^m\) are exchangeable.
\end{theorem}
The proof of Theorem \ref{theo:exchange} is provided in Appendix \ref{appendix:th1}.
This theorem establishes that exchangeability is preserved, thereby validating the application of conformal prediction to HSI classification. 
Specifically, it guarantees that the generated prediction sets will include the true label with a user-specified coverage probability. 
Additionally, we empirically demonstrate this coverage guarantee in our experiments, as detailed in Table \ref{tab:acc}. 
 Building on this theoretical foundation, we propose a conformal procedure that equips any trained HSI classifier with reliable prediction sets, ensuring that these sets contain the true labels with a user-specified probability.

\subsection{Conformal HSI Classification}
For an HSI \(\mathcal{I}\) with a randomly selected subset of known labels, the following steps outline the process for generating conformal prediction sets with a coverage rate of \(1 - \alpha\):
\begin{itemize}
    \item[1.] For each labeled pixel, construct a cuboid and randomly partition these cuboids into two subsets: \(\mathcal{D}_{\text{train}}\), containing \(N_1\) samples, and \(\mathcal{D}_{\text{cal}}\), containing \(N_2\) samples. Similarly, construct a cuboid for each unlabeled pixel to form the set \(\mathcal{D}_{\text{test}}\), which includes \(N_3\) instances.
    \item[2.] Train a HSI classifier \(\hat{\pi}_{\theta}\) using the training subset \(\mathcal{D}_{\text{train}}\).
    \item[3.] Define the score function \(S\) based on the HSI classifier \(\hat{\pi}_{\theta}\) and compute the non-conformity scores \(s_i = S(B_i, y_i)\) for each sample in the calibration set \(\mathcal{D}_{\text{cal}}\).
    \item[4.] Determine the threshold \(\tau\) using \(\{s_i\}_{i=1}^{N_2}\) by solving:
    \begin{equation*}
    \tau = \inf \left\{ s \mid \frac{\left|\{i \mid s_i \leq s\}\right|}{N_2} \geq \frac{\lceil (N_2 + 1)(1 - \alpha) \rceil}{N_2} \right\}.
    \end{equation*}
    \item[5.] For each \(B_i \in \mathcal{D}_{\text{test}}\), derive the prediction set as follows:
    \begin{equation*}
    \mathcal{C}_{1-\alpha}(B_i; \tau) := \{y \in \mathcal{Y} \mid S(B_i, y) \leq \tau\}.
    \end{equation*}
\end{itemize}
Theorem \ref{theo:exchange} indicates that the non-conformity scores of \(\mathcal{D}_{\text{cal}}\) and \(\mathcal{D}_{\text{test}}\) are exchangeable. Thus, according to Theorem \ref{thm:cp}, it follows that the prediction set \(\mathcal{C}_{1-\alpha}(B_i; \tau)\) will include the true label of the central pixel of \(B_i\) with a probability of \(1 - \alpha\). For the non-conformity score function \(S\), we can utilize any well-defined non-conformity score functions, such as APS, RAPS, or SAPS. However, these score functions rely solely on softmax probabilities and label ranks, overlooking the critical spatial information inherent in HSIs.  This observation motivates the exploration of integrating spatial information into the score calculation for enhanced conformal prediction.
\begin{figure*}
\centering
  \includegraphics[width=0.92\textwidth]{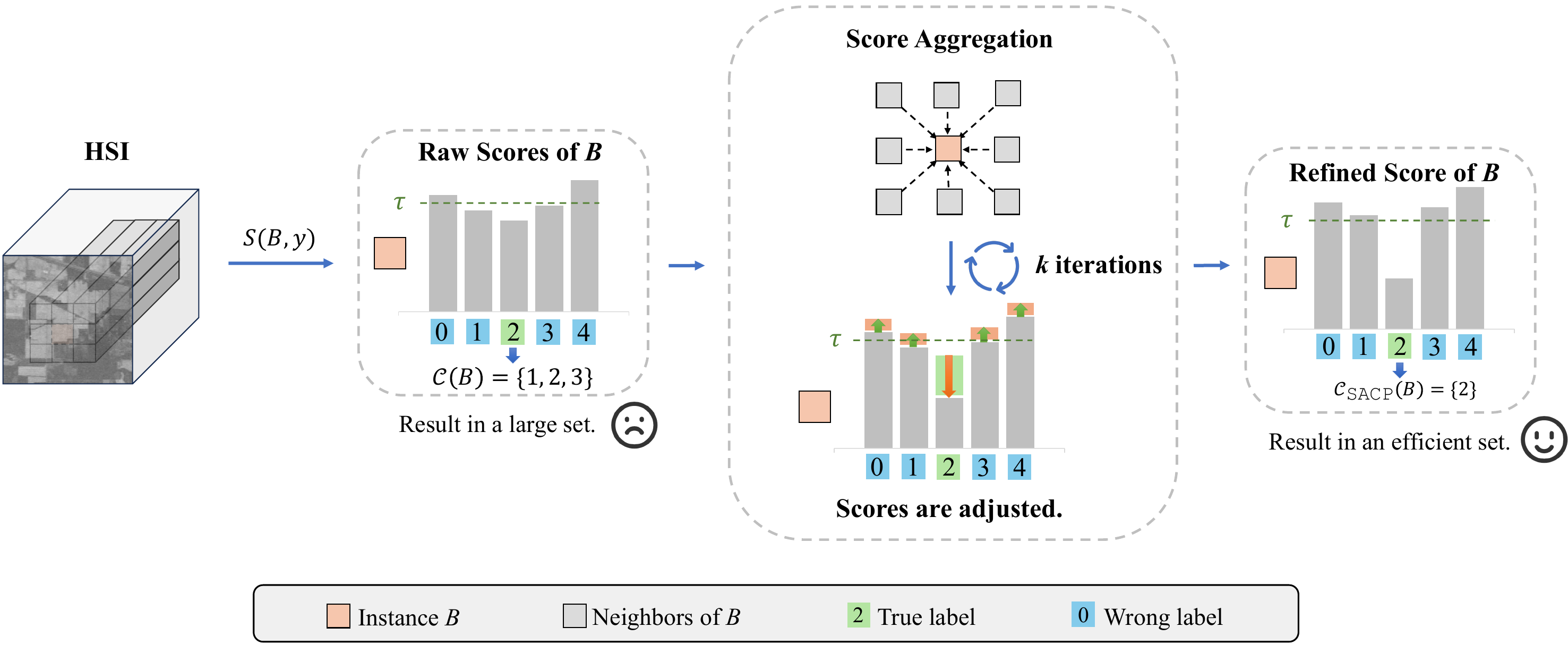}
\caption{\textbf{Overview of \texttt{SACP}}: Raw non-conformity scores lead to excessively large prediction sets. \texttt{SACP} addresses this issue by refining the scores through aggregation of instances with high spatial correlation. This approach is a simple yet effective solution for enhancing the efficiency of prediction sets.}
  \label{fig:main_method}
\end{figure*}

\section{Spatial-Aware Conformal Prediction}
\subsection{Motivation}
Numerous studies have demonstrated that leveraging spatial information can significantly enhance HSI classification accuracy \cite{he2018feature, zhang2023quantum}. However, in the context of conformal prediction, existing non-conformity scores are computed solely based on the output conditional probability of the central pixel and label rankings, resulting in a considerable neglect of spatial information. In the following, we present an theoretical analysis illustrating the impact of incorporating spatial information into the calculation of non-conformity scores.

Given a non-conformity score function \( S \) based on an HSI classifier  \(\hat{\pi}_{\theta}\), the score for instance-label pair \((B_i, y_i)\) is \( S(B_i, y_i) \). To incorporate information from neighboring cuboids, we define the refined score function \(\tilde{S}(B_i, y_i)\) by:
\begin{equation}
\label{eq:aggre}
    \tilde{S}(B_i, y_i) = S(B_i, y_i) + \frac{1}{|\mathcal{N}_i|} \sum_{B_j \in \mathcal{N}_i} S(B_j, y_i),
\end{equation}
where \(\mathcal{N}_i\) denotes the set of neighboring cuboids around the central pixel of \(B_i\). Intuitively speaking, since neighboring pixels in a given HSI often belong to the same class, if \( y_i \) is the true label for \( B_i \), it is likely that \( y_i \) is also the true label for \( B_j \in \mathcal{N}_i \). Consequently, the term \(\frac{1}{|\mathcal{N}_i|} \sum_{B_j \in \mathcal{N}_i} S(B_j, y_i)\) will generally be small. Conversely, if \( y_i \) is not the correct label for \( B_i \), this summation is expected to be larger. Therefore, compared to \(S(B_i, y_i)\), the refined score \(\tilde{S}(B_i, y_i)\) increases modestly for correct instance-label pairs and substantially for incorrect ones. To clarify, the advantages of this aggregation are formally demonstrated in the following proposition.
\begin{proposition}
\label{pro:score}
   Consider two different score functions \(S\) and \(\hat{S}\) based on an HSI classifier. Let \(\mathcal{D}_{\text{cal}}\) represent the calibration set, containing \(n\) samples, and let \(\mathcal{D}_{\text{test}}\) denote the test set, containing \(n'\) instances. We define  \(\hat{\mathcal{D}}_{\text{test}}\) as follows:
    \begin{equation*}
        \hat{\mathcal{D}}_{\text{test}} := \{(B_i, y) \mid (B_i,y_i) \in \mathcal{D}_{\text{test}}, y \in \mathcal{Y}\}.
    \end{equation*}
    Let \(\{s_i\}_{i=1}^{n}\) and \(\{\hat{s}_i\}_{i=1}^{n}\) be the respective scores for \(\mathcal{D}_{\text{cal}}\) based on \( S \) and \( \hat{S} \). The scores for \(\hat{\mathcal{D}}_{\text{test}}\) are denoted by \(\{t_k\}_{k=1}^{n'|\mathcal{Y}|}\) and \(\{\hat{t}_k\}_{k=1}^{n'|\mathcal{Y}|}\), respectively. We define
\[
{R}(\{s_i\}_{i=1}^{n},\{t_k\}_{k=1}^{n'|\mathcal{Y}|}) = \frac{1}{n \times n'|\mathcal{Y}|} \sum_{i=1}^{n} \sum_{k=1}^{n'|\mathcal{Y}|} \mathds{1}_{\{s_i > t_k\}},
\]
which represents the empirical probability that a score from \(\mathcal{D}_{\text{cal}}\) is greater than the one from \(\hat{\mathcal{D}}_{\text{test}}\). Then, we have
\begin{equation*}
\begin{aligned}
     \phantom{~~~~} {R}(\{s_i\}_{i=1}^{n},\{t_k\}_{k=1}^{n'|\mathcal{Y}|}) &> {R}(\{\hat{{s}}_i\}_{i=1}^{n},\{\hat{t}_k\}_{k=1}^{n'|\mathcal{Y}|}) \Leftrightarrow\\
      \sum_{B: (B,y) \in \mathcal{D}_{\text{test}}} \int_0^1 |\mathcal{C}_{1-\alpha}(B)| \, \mathrm{d}\alpha &> \sum_{B: (B,y) \in \mathcal{D}_{\text{test}}} \int_0^1 |\hat{\mathcal{C}}_{1-\alpha}(B)| \, \mathrm{d}\alpha,
\end{aligned}
\end{equation*}
where \(\mathcal{C}(\cdot)\) and \(\hat{\mathcal{C}}(\cdot)\) represent the prediction sets yielded by \(S\) and \(\hat{S}\), and \(\alpha\) denotes the error rate.
\end{proposition}
The proof of Proposition \ref{pro:score} is provided in Appendix \ref{appendix:pro1}.  This proposition suggests that the probability of a randomly selected score from \(\mathcal{D}_{\text{cal}}\) exceeding a randomly selected one from \(\hat{\mathcal{D}}_{\text{test}}\) serves as an indicator of the efficiency of conformal predictors: a lower probability indicates more efficient prediction sets.  For the refined score \(\tilde{S}\) in Eq. \eqref{eq:aggre}, which incorporates aggregation within neighborhoods, scores in \(\hat{\mathcal{D}}_{\text{test}}\) typically show a larger increase than those in \(\mathcal{D}_{\text{cal}}\).   This discrepancy occurs because the instance-label pairs in \(\mathcal{D}_{\text{cal}}\) are always correctly matched, while the defined set \(\hat{\mathcal{D}}_{\text{test}}\) mostly contains mismatched pairs.  Therefore, compared to standard conformal prediction, the integration of spatial information (via Eq. \eqref{eq:aggre}) is expected to lower the probability defined in Proposition \ref{pro:score}, resulting in more efficient prediction sets.

\subsection{Methodology}
In our preceding analysis, we demonstrate that incorporating scores from neighboring pixels potentially enhances the efficiency of prediction sets. To fully leverage the spatial information inherent in HSIs, we introduce Spatial-Aware Conformal Prediction (\texttt{SACP}). Central to this framework is the Score Aggregation Operator \(\mathcal{V}_k\), which is defined as follows:
\begin{equation}
\label{eq:aggrea}
\mathcal{V}_k(B_i,y) = (1-\lambda)\mathcal{V}_{k-1}(B_i,y) + \frac{\lambda}{|\mathcal{N}_i|} \sum_{B_j \in \mathcal{N}_i} \mathcal{V}_{k-1}(B_j,y),
\end{equation}
where \(k\) represents the iteration number, and \(\mathcal{V}_0 = S\) is the utilized base score function. The parameter 
$\lambda$ serves as a trade-off parameter to balance the influence of the neighborhood information. The term \(\mathcal{N}_i\) denotes the set of neighboring cuboids around the central pixel of \(B_i\) that are not in  $\mathcal{D}_{\text{train}}$. For \(k = 1\), each score is aggregated with its immediate neighbors. As \(k\) increases, the scores are computed based on previously aggregated scores, thus incorporating spatial information from progressively larger neighborhoods. An overview of \texttt{SACP} is illustrated in Figure \ref{fig:main_method}. While raw scores often result in excessively large prediction sets, \texttt{SACP} refines these scores by aggregating them according to spatial correlations, thereby yielding more efficient prediction sets. Additionally, the complete pipeline of \texttt{SACP} is detailed in Algorithm \ref{al:SACP}.
\begin{algorithm}
\caption{\textbf{Spatial-Aware Conformal Prediction} (\texttt{SACP})}
\begin{algorithmic}
\STATE \textbf{Input:} Training set \(\mathcal{D}_{\text{train}}\), calibration set \(\mathcal{D}_{\text{cal}}\), test cuboid \(B_{n+1}\), error rate \(\alpha\), and 
parameters \(k\) and \(\lambda\).

    \STATE \textbf{Output:}  Conformal prediction set \(\mathcal{C}_{1-\alpha}(B_{n+1}; \hat{\tau})\).

    \STATE \textbf{Step 1:} Train a deep HSI classifier \(\hat{\pi}_{\theta}\) using \(\mathcal{D}_{\text{train}}\).
    
    \STATE \textbf{Step 2:} Define the score function \(S\) based on \(\hat{\pi}_{\theta}\).
    
    \STATE \textbf{Step 3:} Compute the scores for \(\mathcal{D}_{\text{cal}}\) and aggregate the results using the Score Aggregation Operator defined in  Eq. \eqref{eq:aggrea}. Denote the aggregated scores as \(\{\hat{s}_i\}_{i=1}^{{\scriptstyle|\mathcal{D}_{\text{cal}}|}}\).
    
    \STATE \textbf{Step 4:} Determine the threshold \(\hat{\tau}\)  by:
    \begin{equation*}
    \hat{\tau} = \inf \left\{ s \mid \frac{\left|\{i \mid \hat{s}_i \leq s\}\right|}{N_2} \geq \frac{\lceil (N_2 + 1)(1 - \alpha) \rceil}{N_2} \right\}.
    \end{equation*}
    
    \STATE \textbf{Step 5:} For \(B_{n+1}\),  the prediction set is derived as:
    \begin{equation*}
    \hat{\mathcal{C}}_{1-\alpha}(B_{n+1}; \hat{\tau}) := \{y \in \mathcal{Y} \mid \mathcal{V}_k(B_{n+1}, y) \leq \hat{\tau}\},
    \end{equation*}
    where  $\mathcal{V}_k$ is the Score Aggregation Operator in  Eq. \eqref{eq:aggrea}.
\end{algorithmic}
\label{al:SACP}
\end{algorithm}

To demonstrate that \texttt{SACP} can produce trustworthy prediction sets, we also theoretically establish that \texttt{SACP} ensures coverage, which is  formalized in the following proposition.
\begin{proposition}
\label{pro:sacpcon}
Consider the Score Aggregation Operator \(\mathcal{V}_k\), with \(\mathcal{V}_0 = S\) as the initial score function. Let \(\{(B_i, y_i)\}_{i=1}^n\) be the calibration set and \((B_{n+1}, y_{n+1})\) be a sample from the test set. Denote the aggregated scores for the calibration set by \(\{\hat{s}_i\}_{i=1}^n\), where \(\hat{s}_i = \hat{s}_{i}^{(k)} := \mathcal{V}_k(B_i, y_i)\). Suppose the threshold \(\hat{\tau}\) is computed by:
\begin{equation*}
\hat{\tau} = \inf \left\{ s \mid \frac{\left|\{i \mid \hat{s}_i \leq s\}\right|}{n} \geq \frac{\lceil (n + 1)(1 - \alpha) \rceil}{n} \right\}.
\end{equation*}
Then, the following coverage guarantee holds:
\begin{equation*}
\mathbb{P}\left(y_{n+1} \in \hat{\mathcal{C}}_{1-\alpha}(B_{n+1}; \hat{\tau})\right) \geq 1 - \alpha,
\end{equation*}
where \(\hat{\mathcal{C}}_{1-\alpha}(B_{n+1}; \hat{\tau})\) is formulated as
\[
\hat{\mathcal{C}}_{1-\alpha}(B_{n+1}; \hat{\tau}) := \{y \in \mathcal{Y} \mid \mathcal{V}_k(B_{n+1}, y) \leq \hat{\tau}\}.
\]
\end{proposition}
Proposition \ref{pro:sacpcon} can be proved using mathematical induction, and the proof is provided in Appendix \ref{appendix:pro2}. This proposition formally establishes that \texttt{SACP} maintains the coverage guarantee and can generate prediction sets with the specified coverage probability. Moreover, it offers several compelling advantages:
\begin{itemize}
    \item \textbf{Model-agnostic:} \texttt{SACP} is a post-hoc method that can be incorporated into \textbf{any} pre-trained HSI classifier without necessitating modifications to the model itself.
    \item \textbf{Easy to use:} \texttt{SACP} requires minimal hyperparameter tuning, as it exhibits insensitivity to the iteration count \( k \) and the term weight \( \lambda \) (see Figure \ref{fig:hyper}).
    \item \textbf{Computationally efficient:} \texttt{SACP} maintains computational efficiency comparable to that of standard conformal prediction (see the discussion in Section \ref{dis:compu}).
\end{itemize}

\section{Experiments}
\subsection{Experimental Setup}
\paragraph{Datasets} We utilize the commonly used benchmarks Indian Pines (IP), Pavia University (PU), and Salinas (SA) datasets\footnote{These datasets are available at \url{https://www.ehu.eus/ccwintco/index.php/Hyperspectral_Remote_Sensing_Scenes}.} for our evaluations. Specifically, IP consists of \(145 \times 145\) pixels and 200 valid spectral bands, representing 16 different land-cover classes. PU includes \(610 \times 340\) pixels with 103 usable spectral bands, covering 9 land-cover categories. SA comprises \(512 \times 217\) pixels with 204 effective spectral bands, corresponding to 16 ground-truth classes. For all HSI datasets, we start by randomly sampling HSI cuboids to create the training set. The remaining data is then divided into calibration and test sets according to a pre-defined ratio.
\begin{table}[!t]
\renewcommand{\arraystretch}{0.5}
\caption{Allocation of the three datasets across different classifiers. }
\vspace{10pt}
\centering
\resizebox{0.35\textwidth}{!}{
\setlength{\tabcolsep}{3mm}{ 
\begin{tabular}{cccc} 
\toprule
Model & $\mathcal{D}_{\mathcal{I}}$ & $|\mathcal{D}_{\text{train}}|$ & $|\mathcal{D}_{\text{cal}}|$ + $|\mathcal{D}_{\text{test}}|$\\
\midrule
\multirow{3}{*}[-0.5em]{1D-CNN} & IP  & 250 & 9999\\
\cmidrule(lr){2-4}
 & PU & 103 & 42673\\
\cmidrule(lr){2-4}
 & SA & 244 & 53885\\
\cmidrule(lr){1-4}
\multirow{3}{*}[-0.5em]{3D-CNN} & IP & 128 & 10113\\
\cmidrule(lr){2-4}
 & PU & 52 & 42724\\
\cmidrule(lr){2-4}
 & SA & 122 & 54007\\
\cmidrule(lr){1-4}
\multirow{3}{*}[-0.5em]{HybridSN} & IP & 128 & 10113\\
\cmidrule(lr){2-4}
 & PU & 52 & 42724\\
\cmidrule(lr){2-4}
 & SA & 122 & 54007\\
\cmidrule(lr){1-4}
\multirow{3}{*}[-0.5em]{SSTN} & IP & 128 & 10113\\
\cmidrule(lr){2-4}
 & PU & 52 & 42724\\
\cmidrule(lr){2-4}
 & SA & 122 & 54007\\
\bottomrule
\end{tabular}
}}
\label{tab:patch}
\end{table}
\begin{table}[!t]
\renewcommand{\arraystretch}{1.5}
\caption{Performance of the pre-trained HSI classifiers used in our experiments.}
\centering
\resizebox{0.45\textwidth}{!}{
\setlength{\tabcolsep}{2mm}{ 
\begin{tabular}{c@{\hskip 10pt}ccccc} 
\toprule
$\mathcal{D}_{\mathcal{I}}$ & Metrics & 1D-CNN & 3D-CNN & HybridSN & SSTN\\
\midrule
\multirow{2}{*}{IP} & OA (\%)  & 68.44 & 68.93 & 75.25 & 88.35\\
& AA (\%)  & 64.38 & 69.10 & 74.53 & 87.43\\
\cmidrule(lr){1-6}
\multirow{2}{*}{PU} & OA (\%)  & 70.45 & 68.97 & 68.31 & 89.39\\
& AA (\%)  & 67.88 & 65.81 & 51.82 & 88.26\\
\cmidrule(lr){1-6}
\multirow{2}{*}{SA} & OA (\%)  & 86.44 & 86.11 & 87.91 & 88.52\\
& AA (\%)  & 85.20 & 87.61 & 87.35 & 88.61\\
\bottomrule
\end{tabular}
}}
\label{tab:acc}
\end{table}
  
\paragraph{Pre-trained HSI classifiers} 
To validate the generalizability of standard conformal prediction and \texttt{SACP} in HSI classification, we employ four widely used HSI classifiers, including both CNN-based and Transformer-based models, each featuring distinct scales and architectures. 
For CNN-based models, we select 1D-CNN \cite{hu2015deep}, 3D-CNN \cite{hamida20183}, and HybridSN \cite{roy2019hybridsn}. The 1D-CNN and 3D-CNN models differ in their convolutional kernel dimensions, while HybridSN leverages multi-dimensional kernels to improve feature extraction. 
As for the Transformer-based model, we employ SSTN \cite{zhong2021spectral}.
Each HSI classifier is initially pre-trained on the three datasets and subsequently evaluated on the corresponding calibration and test data. 
We follow the SSTN training framework available on GitHub\footnote{\url{https://github.com/zilongzhong/SSTN}} for pre-training these classifiers. Table \ref{tab:patch} details the sizes of the training, calibration, and test sets for each dataset across the models. By default, we set the calibration and test sets to be of equal size. Notably, our method remains stable across varying calibration set sizes (see Table \ref{tab:change}).

\begin{table*}[!t]

\renewcommand{\arraystretch}{0.9}
\caption{Performance comparison of prediction sets generated by \texttt{SACP} and standard conformal prediction (\texttt{SCP}) across various HSI classifiers  with different error rates \(\alpha\). \textbf{Bold} numbers  indicate that \texttt{SACP} is superior.  \(\downarrow\) indicates that smaller values are better.}
\centering
\resizebox{0.974\textwidth}{!}{
\setlength{\tabcolsep}{3.2mm}{ 
\begin{tabular}{@{}cc@{\hskip 10pt}c@{\hskip 10pt}cccccc@{\hskip 7pt}} 
\toprule
\multirow{3}*[-5pt]{$\mathbf{\mathcal{D}_{\mathcal{I}}}$}& \multirow{3}*[-5pt]{Model}&  \multirow{3}*[-5pt]{Score} & \multicolumn{6}{c}{w/ \texttt{SCP} $\backslash$ w/ \texttt{SACP}} \\
\cmidrule(lr){4-9}
  & & &\multicolumn{3}{c}{{$\alpha = 0.05$}} & \multicolumn{3}{c}{{$\alpha = 0.1$}}\\
 \cmidrule(lr){4-6} \cmidrule(lr){7-9}
  & & & Coverage & Size ($\downarrow$) & SSCV ($\downarrow$)  & Coverage & Size ($\downarrow$) & SSCV ($\downarrow$) \\
\midrule
\multirow{12}{*}[-1em]{\rotatebox[origin=c]{90}{Indian Pines}} 
& \multirow{3}{*}[-1pt]{1D-CNN}  & APS &0.95 $\backslash$ 0.95 &3.68 $\backslash$ \textbf{2.28} & 0.41 $\backslash$ \textbf{0.28} &0.90 $\backslash$ 0.90 & 2.52 $\backslash$ \textbf{1.75} & 0.51 $\backslash$ \textbf{0.45}\\
& & RAPS & 0.95 $\backslash$ 0.94 & 4.09 $\backslash$ \textbf{2.29} & 0.54 $\backslash$ 0.57 &0.90 $\backslash$ 0.90 & 2.54 $\backslash$ \textbf{1.74} & 0.88 $\backslash$ \textbf{0.30}\\
& & SAPS & 0.95 $\backslash$ 0.95 & 6.68 $\backslash$ \textbf{4.31} & 2.48 $\backslash$ \textbf{1.83} &0.90 $\backslash$ 0.90 & 5.56 $\backslash$ \textbf{3.02} & 4.84 $\backslash$ \textbf{1.95}\\
\cmidrule(lr){2-9}
& \multirow{3}{*}[-1pt]{3D-CNN}  & APS &0.94 $\backslash$ 0.95 & 5.73 $\backslash$ \textbf{3.27} & 0.47 $\backslash$ \textbf{0.38} &0.90 $\backslash$ 0.90 & 2.85 $\backslash$ \textbf{2.06} & 0.35 $\backslash$ \textbf{0.34}\\
& & RAPS &0.95 $\backslash$ 0.95 & 4.12 $\backslash$ \textbf{3.92} & 0.10 $\backslash$ 0.21 &0.90 $\backslash$ 0.90 & 3.21 $\backslash$ \textbf{2.38} & 0.11 $\backslash$ 0.25\\
& & SAPS &0.95 $\backslash$ 0.95 & 6.40 $\backslash$ \textbf{5.44} & 0.98 $\backslash$ \textbf{0.95} &0.91 $\backslash$ 0.90 & 4.97 $\backslash$ \textbf{3.72} & 1.61 $\backslash$ \textbf{1.51}\\
\cmidrule(lr){2-9}
& \multirow{3}{*}[-1pt]{HybridSN}  & APS &0.95 $\backslash$ 0.95 & 5.56 $\backslash$ \textbf{4.83} & 0.20 $\backslash$ \textbf{0.16} &0.90 $\backslash$ 0.90 & 3.28 $\backslash$ \textbf{2.74} & 0.92 $\backslash$ \textbf{0.15}\\
& & RAPS &0.95 $\backslash$ 0.95 & 7.34 $\backslash$ \textbf{7.12} & 0.40 $\backslash$ 0.96 &0.90 $\backslash$ 0.90 &  4.29 $\backslash$ \textbf{3.95} & 0.98 $\backslash$ \textbf{0.75}\\
& & SAPS &0.95 $\backslash$ 0.95 & 6.79 $\backslash$ \textbf{6.07} & 0.19 $\backslash$ 0.41 &0.90 $\backslash$ 0.90 &  4.03 $\backslash$ \textbf{3.40} & 0.64 $\backslash$ \textbf{0.61}\\
\cmidrule(lr){2-9}
& \multirow{3}{*}[-1pt]{SSTN}  &APS & 0.95 $\backslash$ 0.95& 2.81 $\backslash$ \textbf{1.73} & 0.42 $\backslash$ \textbf{0.18} &0.90 $\backslash$ 0.90 & 2.00 $\backslash$ \textbf{1.38} & 0.41 $\backslash$ 0.47\\
& & RAPS &0.95 $\backslash$ 0.95 & 2.52 $\backslash$ \textbf{1.62} & 0.29 $\backslash$ 0.30 & 0.90 $\backslash$ 0.90& 1.87 $\backslash$ \textbf{1.36} & 0.29 $\backslash$ 0.34\\
& & SAPS &0.95 $\backslash$ 0.95 & 6.98 $\backslash$ \textbf{4.16} & 4.38 $\backslash$ \textbf{1.74} & 0.90 $\backslash$ 0.90& 5.33 $\backslash$ \textbf{3.25} & 6.35 $\backslash$ \textbf{3.09}\\
\midrule
\multirow{12}{*}[-1em]{\rotatebox[origin=c]{90}{Pavia University}} & \multirow{3}{*}[-1pt]{1D-CNN}  & APS &0.95 $\backslash$ 0.95 &2.26 $\backslash$ \textbf{1.92} & 0.39 $\backslash$ \textbf{0.37} &0.90 $\backslash$ 0.90 & 1.65 $\backslash$ \textbf{1.57} & 0.40 $\backslash$ \textbf{0.27}\\
& & RAPS & 0.95 $\backslash$ 0.95 & 2.00 $\backslash$ \textbf{1.83} & 0.23 $\backslash$ 0.31 &0.90 $\backslash$ 0.90 & 1.59 $\backslash$ \textbf{1.54} & 0.40 $\backslash$ \textbf{0.29}\\
& & SAPS & 0.95 $\backslash$ 0.95 & 3.92 $\backslash$ 3.99 & 1.77 $\backslash$ 2.21 &0.90 $\backslash$ 0.90 & 3.44 $\backslash$ \textbf{3.04} & 3.62 $\backslash$ \textbf{2.68}\\
\cmidrule(lr){2-9}
&\multirow{3}{*}[-1pt]{3D-CNN}  & APS &0.94 $\backslash$ 0.95 & 2.77 $\backslash$ \textbf{2.34} & 1.04 $\backslash$ \textbf{0.69} &0.89 $\backslash$ 0.89 & 2.14 $\backslash$ \textbf{1.79} & 0.94 $\backslash$ \textbf{0.66}\\
& & RAPS &0.94 $\backslash$ 0.94 & 2.57 $\backslash$ \textbf{2.28} & 0.41 $\backslash$ 0.46 &0.89 $\backslash$ 0.89 & 2.04 $\backslash$ \textbf{1.76} & 0.64 $\backslash$ \textbf{0.56}\\
& & SAPS &0.94 $\backslash$ 0.94 & 4.80 $\backslash$ \textbf{4.31} & 4.56 $\backslash$ \textbf{3.85} & 0.89 $\backslash$ 0.89 & 4.04 $\backslash$ \textbf{3.32} & 6.56 $\backslash$ \textbf{4.23}\\
\cmidrule(lr){2-9}
& \multirow{3}{*}[-1pt]{HybridSN}  & APS &0.95 $\backslash$ 0.95 & 4.79 $\backslash$ \textbf{4.59} & 4.59 $\backslash$ \textbf{3.58} &0.90 $\backslash$ 0.90 & 3.38 $\backslash$ \textbf{3.01} & 2.39 $\backslash$ \textbf{1.33}\\
& & RAPS &0.94 $\backslash$ 0.95 & 5.50 $\backslash$ \textbf{5.36} & 0.47 $\backslash$ \textbf{0.36} &0.89 $\backslash$ 0.90 &  3.70 $\backslash$ 3.74 & 0.81 $\backslash$ \textbf{0.07}\\
& & SAPS &0.95 $\backslash$ 0.95 & 5.57 $\backslash$ \textbf{5.44} & 1.98 $\backslash$ 2.35 &0.90 $\backslash$ 0.90 &  3.99 $\backslash$ \textbf{3.71} & 3.33 $\backslash$ 3.47\\
\cmidrule(lr){2-9}
& \multirow{3}{*}[-1pt]{SSTN}  &APS & 0.95 $\backslash$ 0.95& 1.75 $\backslash$ \textbf{1.24} & 0.22 $\backslash$ 0.26 &0.90 $\backslash$ 0.90 & 1.39 $\backslash$ \textbf{1.11} & 0.23 $\backslash$ 0.29\\
& & RAPS &0.95 $\backslash$ 0.95 & 1.60 $\backslash$ \textbf{1.22} & 0.20 $\backslash$ 0.20& 0.90 $\backslash$ 0.90& 1.33 $\backslash$ \textbf{1.10} & 0.13 $\backslash$ 0.23\\
& & SAPS &0.95 $\backslash$ 0.95 & 3.26 $\backslash$ \textbf{2.24} & 2.07 $\backslash$ \textbf{0.92}& 0.90 $\backslash$ 0.90& 2.75 $\backslash$ \textbf{1.91} & 2.96 $\backslash$ \textbf{1.64}\\
\midrule
\multirow{12}{*}[-1em]{\rotatebox[origin=c]{90}{Salinas}} & \multirow{3}{*}[-1pt]{1D-CNN}  & APS &0.95 $\backslash$ 0.95 & 1.40 $\backslash$ \textbf{1.20} & 0.15 $\backslash$ 0.15 &0.90 $\backslash$ 0.90 & 1.20 $\backslash$ \textbf{1.09} & 0.18 $\backslash$ 0.25\\
& & RAPS & 0.95 $\backslash$ 0.95 & 1.37 $\backslash$ \textbf{1.20} & 0.06 $\backslash$ 0.20 &0.90 $\backslash$ 0.90 & 1.19 $\backslash$ \textbf{1.07} & 0.23 $\backslash$ 0.25\\ 
& & SAPS & 0.95 $\backslash$ 0.95 & 3.63 $\backslash$ \textbf{1.71} & 1.20 $\backslash$ \textbf{0.16} &0.90 $\backslash$ 0.90 & 2.97 $\backslash$ \textbf{1.28} & 2.04 $\backslash$ \textbf{0.16}\\ 
\cmidrule(lr){2-9}
& \multirow{3}{*}[-1pt]{3D-CNN}  & APS &0.95 $\backslash$ 0.95 & 1.48 $\backslash$ \textbf{1.25} & 0.20 $\backslash$ \textbf{0.19} &0.90 $\backslash$ 0.90 & 1.17 $\backslash$ \textbf{1.08} & 0.12 $\backslash$ 0.16\\
& & RAPS &0.95 $\backslash$ 0.95 & 1.47 $\backslash$ \textbf{1.24} & 0.13 $\backslash$ 0.18 &0.90 $\backslash$ 0.90 & 1.15 $\backslash$ \textbf{1.07} & 0.15 $\backslash$ 0.17\\
& & SAPS &0.95 $\backslash$ 0.95 & 2.94 $\backslash$ \textbf{2.02} & 0.58 $\backslash$ \textbf{0.28} &0.90 $\backslash$ 0.90 & 2.39 $\backslash$ \textbf{1.29} & 1.07 $\backslash$ \textbf{0.13}\\
\cmidrule(lr){2-9}
& \multirow{3}{*}[-1pt]{HybridSN}  & APS &0.95 $\backslash$ 0.95 & 1.37 $\backslash$ \textbf{1.09} & 0.18 $\backslash$ \textbf{0.07} &0.90 $\backslash$ 0.90 & 1.10 $\backslash$ \textbf{1.03} & 0.18 $\backslash$ 0.23\\
& & RAPS &0.95 $\backslash$ 0.95 & 1.20 $\backslash$ \textbf{1.07} & 0.12 $\backslash$ \textbf{0.10} &0.90 $\backslash$ 0.90 & 1.06 $\backslash$ \textbf{1.00} & 0.12 $\backslash$ 0.31\\
& & SAPS &0.95 $\backslash$ 0.95 & 1.90 $\backslash$ \textbf{1.37} & 0.42 $\backslash$ \textbf{0.33} &0.90 $\backslash$ 0.90 & 1.66 $\backslash$ \textbf{1.18} & 0.69 $\backslash$ \textbf{0.31}\\
\cmidrule(lr){2-9}
&\multirow{3}{*}[-1pt]{SSTN}  &APS & 0.95 $\backslash$ 0.95& 1.70 $\backslash$ \textbf{1.19} & 0.21 $\backslash$ \textbf{0.16} &0.90 $\backslash$ 0.90 & 1.37 $\backslash$ \textbf{1.08} & 0.23 $\backslash$ \textbf{0.10}\\
& & RAPS &0.95 $\backslash$ 0.95 & 1.60 $\backslash$ \textbf{1.18} & 0.12 $\backslash$ 0.14 & 0.90 $\backslash$ 0.90& 1.29 $\backslash$ \textbf{1.06} & 0.10 $\backslash$ 0.11\\
& & SAPS &0.95 $\backslash$ 0.95 & 5.42 $\backslash$ \textbf{2.65} & 3.58 $\backslash$ \textbf{0.86} & 0.90 $\backslash$ 0.90& 3.91 $\backslash$ \textbf{2.07} & 3.64 $\backslash$ \textbf{1.26}\\
\bottomrule
\end{tabular}
}}
\label{tab:size}
\end{table*}

Due to its relatively poor performance, the 1D-CNN is allocated a larger amount of training data compared to the other three models. For clarity, Table \ref{tab:acc} demonstrates the Overall Accuracy (OA) and Average Accuracy (AA) results for the pre-trained HSI classifiers across the three datasets. The definitions of OA and AA are given as follows:
\begin{equation}
    \text{OA}=\frac{1}{|\mathcal{D}_{\text{test}}|}\sum\limits_{i=1}^{|\mathcal{D}_{\text{test}}|}\mathds{1}_{\{\hat{y}_i=y_i\}},
\end{equation}
\begin{equation}
    \text{AA}=\frac{1}{|\mathcal{Y}|}\sum\limits_{y_j\in\mathcal{Y}}\frac{\sum_{i=1}^{|\mathcal{D}_{\text{test}}|}\mathds{1}_{\{\hat{y}_i=y_i, y_i=y_j\}}}{\sum_{i=1}^{|\mathcal{D}_{\text{test}}|}\mathds{1}_{\{y_i=y_j\}}},
\end{equation}
where $\mathcal{D}_{\text{test}}:=\{(B_i,y_i)\}_{i=1}^{|\mathcal{D}_{\text{test}}|}$, \(\hat{y}_i = \mathrm{argmax}(\hat{\pi}_{\theta}(B_i))\), $\mathds{1}$ is the indicator function, and $j\in\{1,\dots,|\mathcal{Y}|\}$. \textbf{Note}: In HSI classifier pre-training, the primary focus is not on maximizing their performance metrics such as OA and AA, but rather on \textbf{achieving model diversity}. Our goal is to validate the effectiveness—i.e., the generalizability—of our approach across different model architectures and training processes, irrespective of their performance levels. Given that our primary goal is not to optimize classification performance, we utilize only a minimal amount of data for pre-training these classifiers.

\paragraph{Implementation Details}
The training configurations for all four HSI classifiers adhere to their recommended settings. Each model is trained for 200 epochs utilizing the Adam optimizer \cite{kingma2014adam}, with a batch size of 128 and a learning rate of 0.002. For \texttt{SACP}, the trade-off parameter \(\lambda\) is \textbf{fixed} at 0.5, and the iteration count \(k\) is set to 1 for all experiments. The primary evaluation metrics are Coverage, Size, and SSCV, as detailed in Section \ref{back:cp}. To enhance statistical stability, each experiment is repeated 30 times, and the results are averaged for reporting. All experiments are performed on an NVIDIA GeForce RTX 3090 using PyTorch \cite{paszke2019pytorch}.

\begin{figure*}
\centering
  \includegraphics[width=0.855\textwidth]{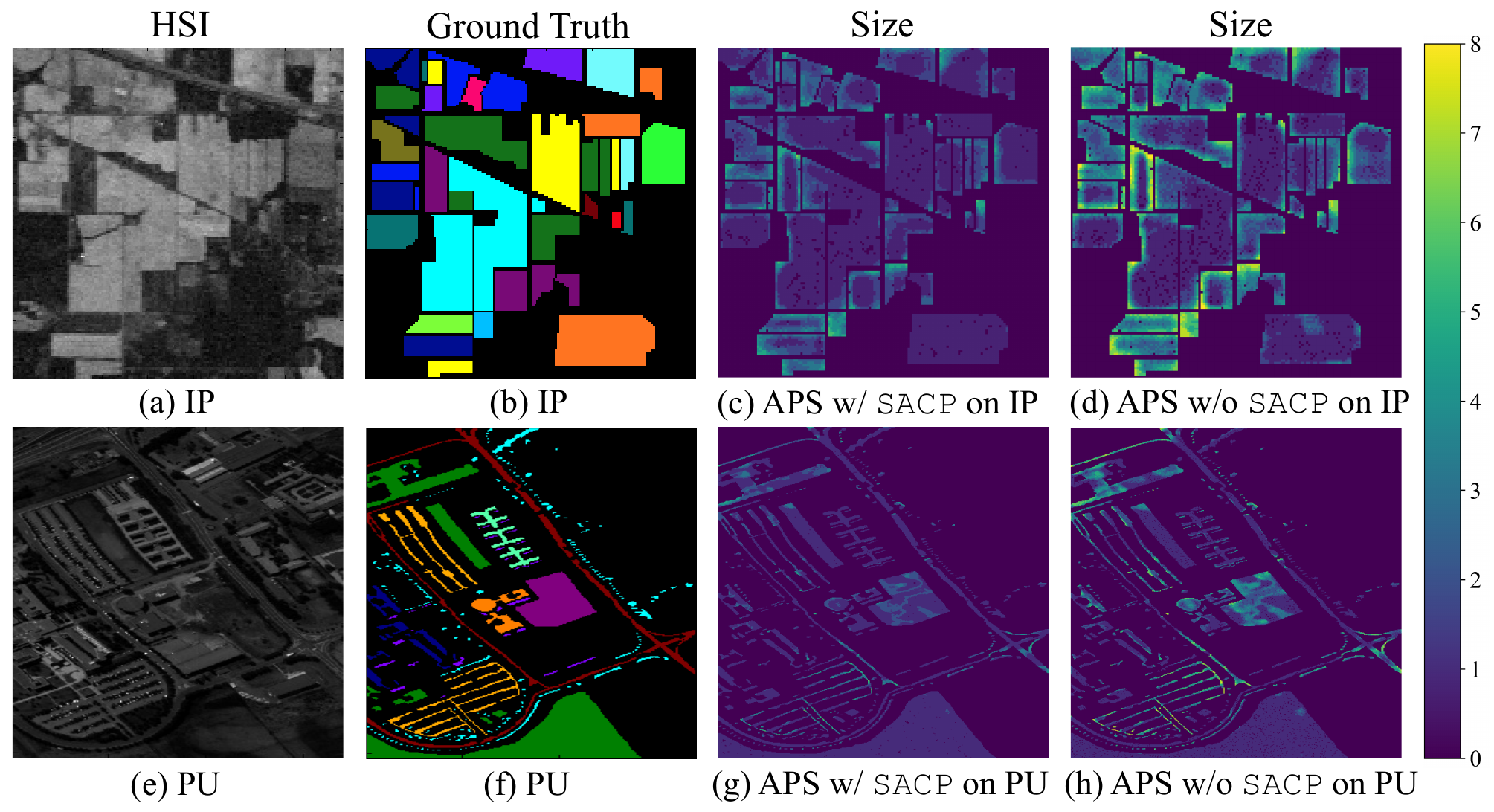}
\caption{Visualizations of the average size of prediction sets for each pixel in \(\mathcal{D}_{\text{cal}}\) and \(\mathcal{D}_{\text{test}}\) are presented for the (a) IP and (e) PU datasets. The true labels for each class are shown in (b) and (f). The brightness of each pixel indicates the size of its prediction set, with brighter pixels representing larger sets. Panels (c) and (g) display the set sizes with \texttt{SACP}, while panels (d) and (h) show the set sizes with standard conformal prediction.}
  \label{fig:main_visual}
\end{figure*}
\begin{figure}
\centering
\subfloat[1][Effect of $\lambda$ on Size of \texttt{SACP}.]
{\label{subfig:lmd}\includegraphics[scale=0.18]{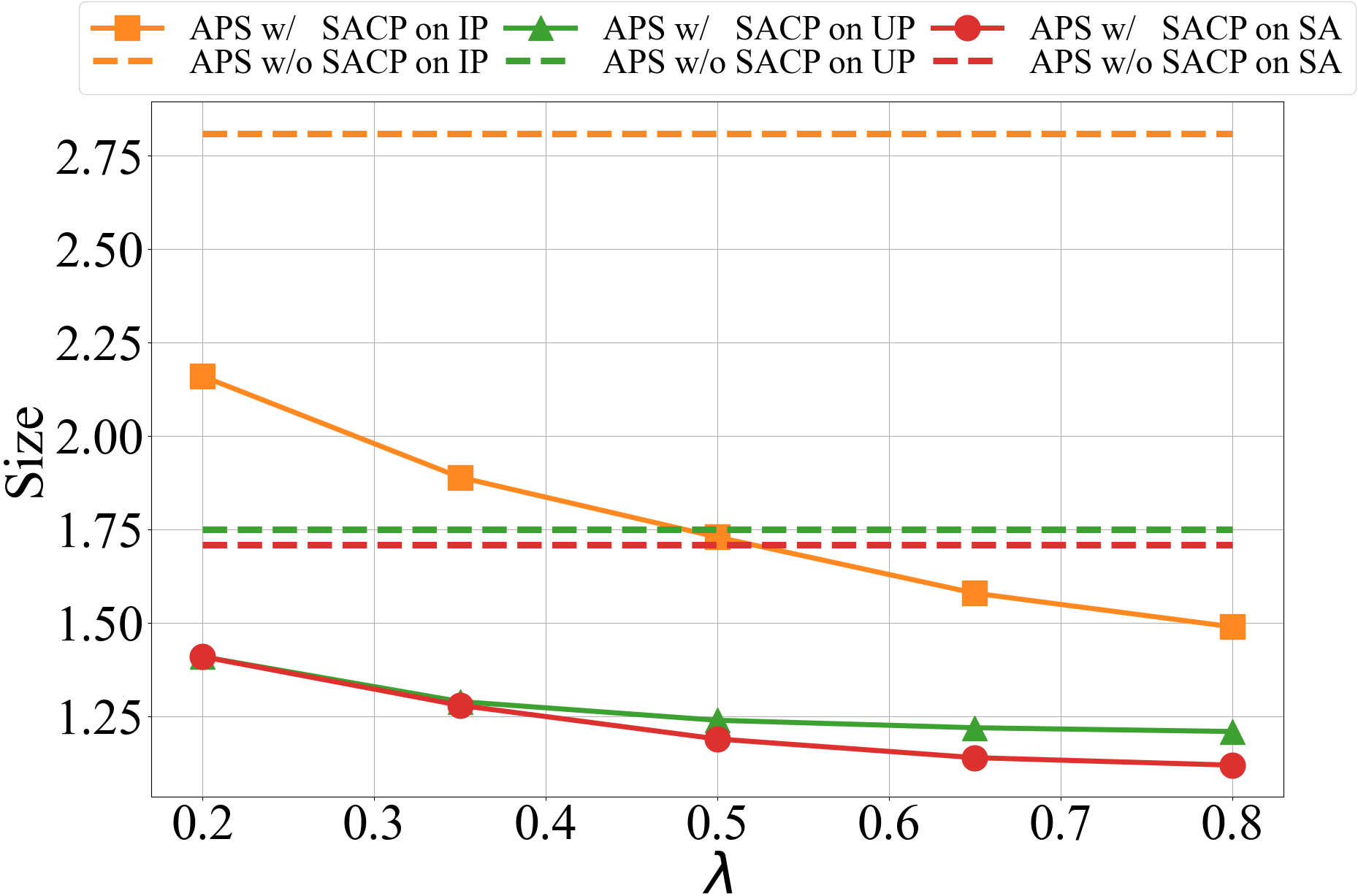}}\\
\subfloat[2][Effect of $k$ on Size of \texttt{SACP}.]
{\label{subfig:n}\includegraphics[scale=0.18]{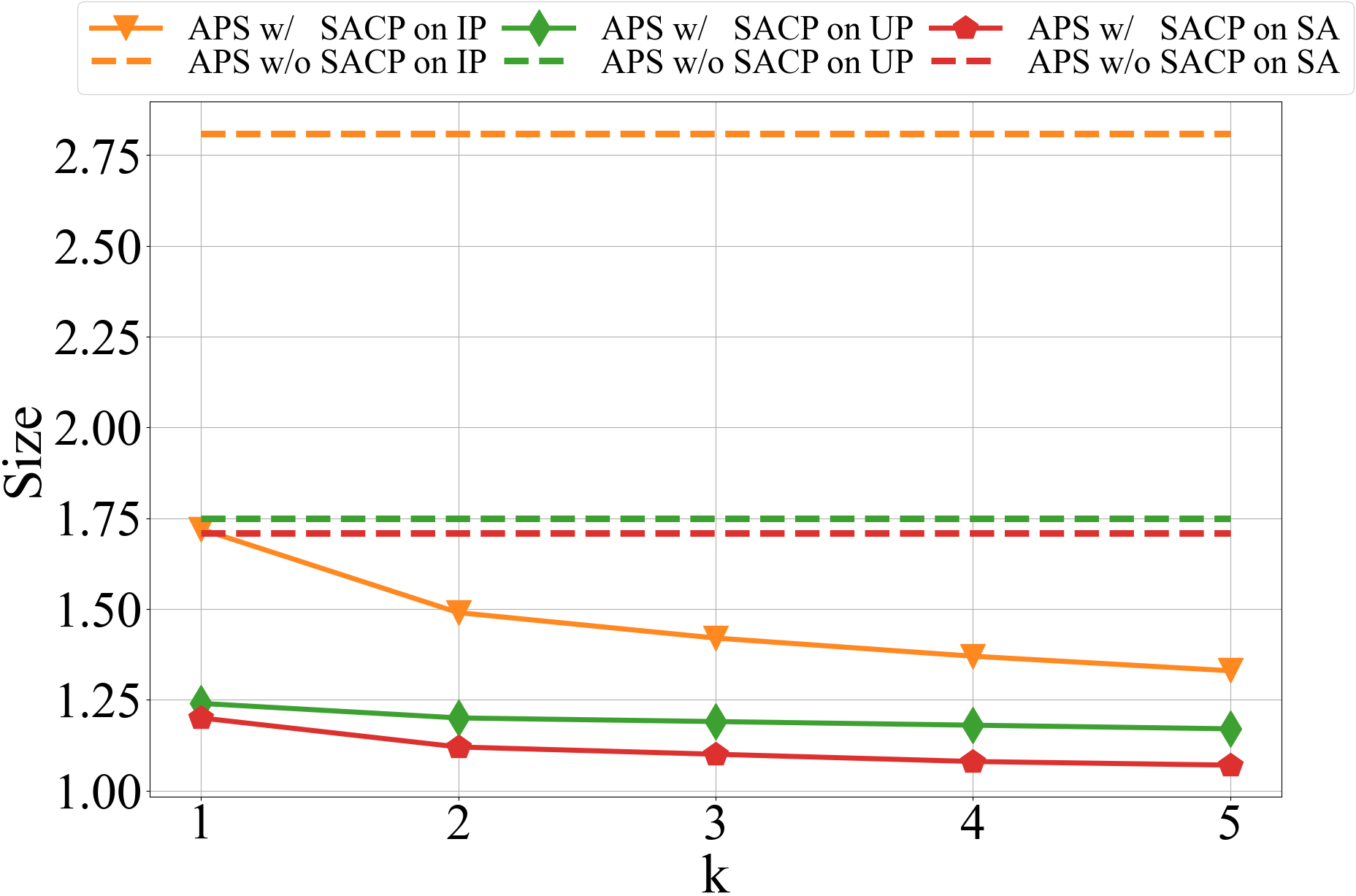}}
\caption{\textbf{Effect of $\lambda$ and $k$ on Prediction Set Size:} Each subplot presents results from three comparative experiments conducted on different datasets. The dashed line indicates the performance of APS without \texttt{SACP}.}
\label{fig:hyper}
\end{figure}

\subsection{Main Results}
\paragraph{\texttt{SACP} generates more efficient prediction sets} We compare the performance of prediction sets generated by standard conformal prediction and \texttt{SACP} across various HSI classifiers and datasets with different error rates \(\alpha\). 
The results are presented in Table \ref{tab:size}, showing that \texttt{SACP} consistently maintains the desired coverage of \(1 - \alpha\). 
The results empirically validates the coverage guarantee of \texttt{SACP}, as outlined in Proposition \ref{pro:sacpcon}. 
In other words, for each pixel, its true label will be reliably included in the prediction set with the user-specified probabilities.
Additionally, \texttt{SACP} consistently produces more efficient prediction sets while maintaining a satisfactory conditional coverage rate. 
For instance, when applying 3D-CNN to the IP dataset, \texttt{SACP} reduces the size from 5.73 for APS to 3.27, representing a 42.9\% decrease. 
Similarly, for RAPS on the same dataset with 1D-CNN, \texttt{SACP} lowers the size from 3.64 to 2.26, reflecting a 37.9\% reduction. 
These improvements are observed across different error rates $\alpha$, non-conformity score functions, model architectures, and datasets, highlighting the robustness and generalizability of \texttt{SACP}. 
Notably, unlike traditional classification tasks, SAPS and RAPS do not consistently outperform APS in the context of HSI classification, underscoring the challenges associated with conformal prediction in this field. Nevertheless, \texttt{SACP} consistently improves the performance of various non-conformity score functions, demonstrating its effectiveness.

\paragraph{\texttt{SACP} consistently reduces prediction set size across various pixels} To further demonstrate the effectiveness of \texttt{SACP}, we visualize the size of prediction sets for each pixel of \(\mathcal{D}_{\text{cal}}\) and \(\mathcal{D}_{\text{test}}\) in Figure \ref{fig:main_visual}. This experiment is conducted on the IP and PU datasets using SSTN, with APS as the base non-conformity score function and \(\alpha\) set to 0.05. In the visualization, brighter pixels represent larger prediction sets. The results clearly indicate that \texttt{SACP} significantly reduces the size of prediction sets compared to standard conformal prediction.  Moreover, standard conformal prediction often produces unusually large prediction sets, even for interior pixels, as evidenced in the lower-left corner of Figure \ref{fig:main_visual}\textcolor{blue}{(d)}. Typically, edge pixels, which are more challenging to predict, tend to have larger prediction sets, while interior pixels have smaller sets. \texttt{SACP} effectively addresses this issue by consistently reducing prediction set sizes for both edge and interior pixels. The experimental results further illustrate the superior performance of our approach in conformal HSI classification.

\begin{table*}[!t]
\renewcommand{\arraystretch}{1.5}
\caption{\textbf{Effect of Calibration Set Size on \texttt{SACP}:} The ratio of calibration samples is given by \(\gamma = \frac{|\mathcal{D}_{\text{cal}}|}{|\mathcal{D}_{\text{cal}} \cup \mathcal{D}_{\text{test}}|}\). }
\centering
\resizebox{0.98\textwidth}{!}{
\setlength{\tabcolsep}{1.5mm}{ 
\begin{tabular}{ccccccccccccc} 
\toprule
$\mathcal{D}_{\mathcal{I}}$&& $\gamma = 0.1$& $\gamma = 0.15$& $\gamma = 0.2$& $\gamma = 0.25$& $\gamma = 0.3$& $\gamma = 0.35$& $\gamma = 0.4$& $\gamma = 0.45$& $\gamma = 0.5$& $\gamma = 0.55$& $\gamma = 0.6$\\
\midrule
\multirow{2}{*}{IP}& Coverage& 0.96 &0.95& 0.95& 0.95 &0.95& 0.95& 0.95& 0.95& 0.95& 0.95& 0.95\\
& Size& 1.83 &1.70&1.73&1.75&1.74&1.73 &1.75&1.73&1.73&1.76&1.72\\
\midrule
\multirow{2}{*}{PU}& Coverage& 0.95 &0.95& 0.95& 0.95 &0.95& 0.95& 0.95& 0.95& 0.95& 0.95& 0.95\\
& Size& 1.24 &1.25 &1.24 &1.24& 1.26& 1.25 & 1.25 & 1.24 & 1.24 & 1.25 &1.24\\
\midrule
\multirow{2}{*}{SA}& Coverage& 0.96 &0.95& 0.95& 0.95 &0.95& 0.95& 0.95& 0.95& 0.95& 0.95& 0.95\\
& Size& 1.23 &1.20 &1.21 &1.19&1.19& 1.20& 1.20 & 1.19 &1.19 & 1.20 & 1.19\\
\bottomrule
\end{tabular}
}}
\label{tab:change}
\end{table*}

\begin{figure}
\centering
\subfloat[1][PU]
{\label{subfig:pudiff}\includegraphics[scale=0.18]{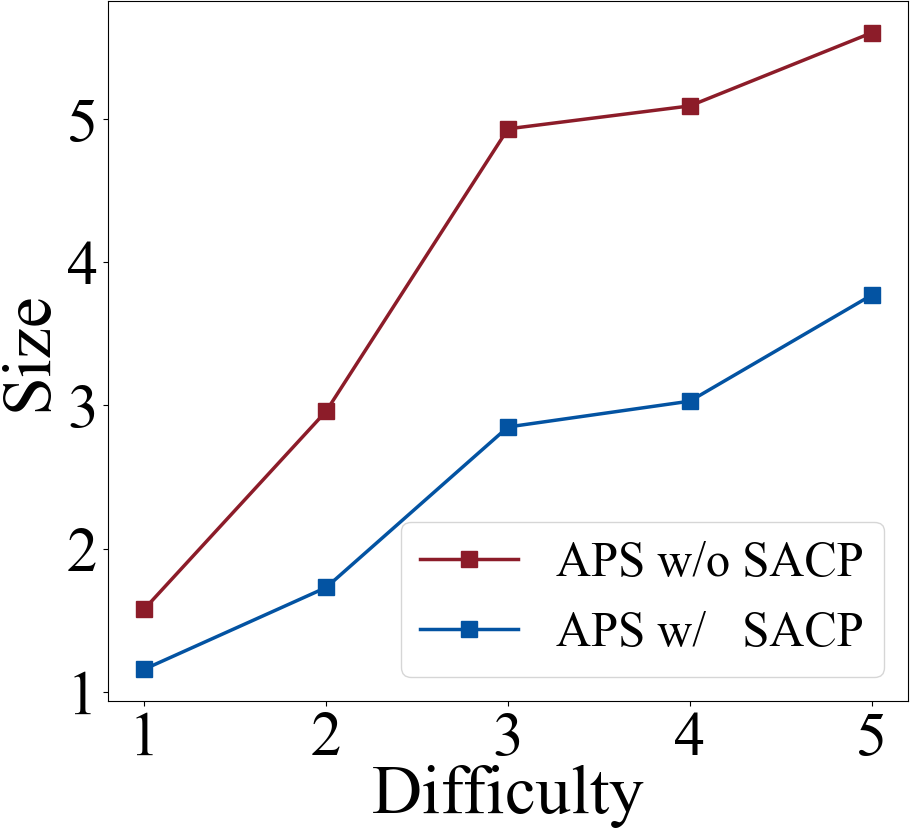}}
\hspace{-1mm}
\subfloat[2][SA]
{\label{subfig:sadiff}\includegraphics[scale=0.18]{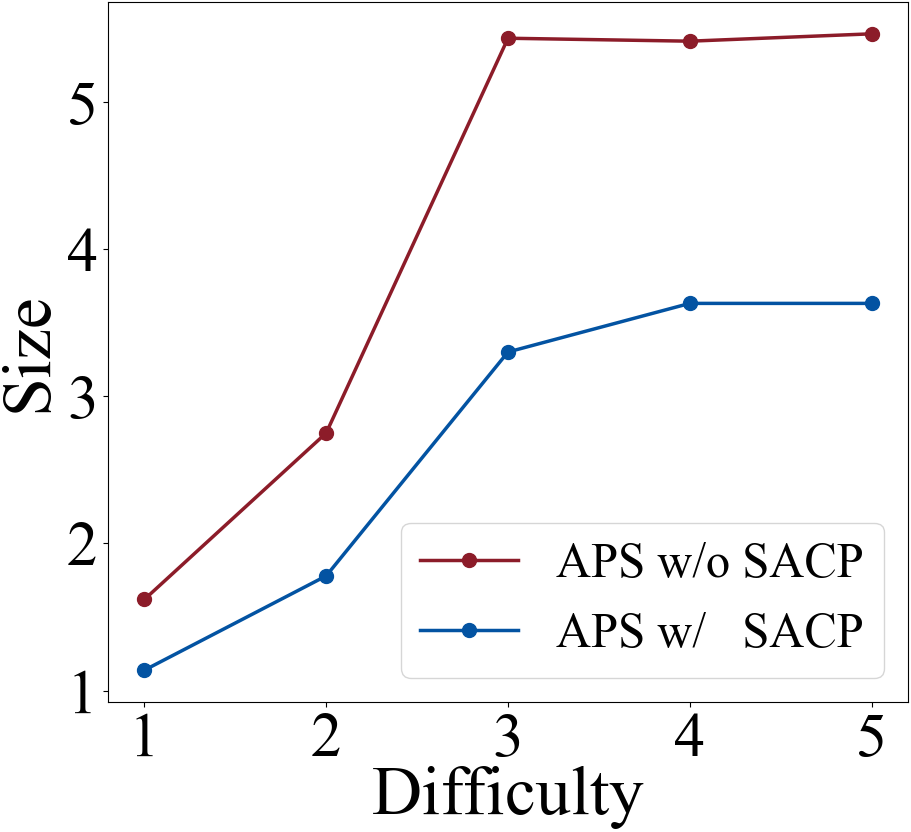}}
\caption{\textbf{Size vs. Difficulty}:  The term ``Difficulty" refers to the ranking of the true class label in the output probability, reflecting the difficulty of accurately classifying the pixel. We report the average size of prediction sets for the PU and SA datasets using SSTN, categorized by their Difficulty.}
\label{fig:sizevsdiff}
\end{figure}

\paragraph{\texttt{SACP} is insensitive to hyperparameters} 
In \texttt{SACP}, two hyperparameters are crucial: \(\lambda\) and \(k\). The parameter \(\lambda\) serves as a trade-off, balancing the influence of the neighborhood, while \(k\) specifies the number of iterations. Increasing \(k\) allows \texttt{SACP} to incorporate spatial information from progressively larger neighborhoods. This section examines the impact of variations in \(\lambda\) and \(k\) on the performance of \texttt{SACP}.  The experiment is conducted on all selected datasets using SSTN, with APS as the score function and \(\alpha\) set to 0.05. The results are illustrated in Figure \ref{fig:hyper}. Figure \ref{subfig:lmd} demonstrates that \texttt{SACP} consistently outperforms the baseline across different values of \(\lambda\), with the size of prediction sets decreasing as \(\lambda\) increases.  Figure \ref{subfig:n} illustrates that \texttt{SACP} consistently achieves superior performance as the number of iterations \(k\) changes. Overall, the results suggest that \texttt{SACP} is relatively insensitive to variations in \(\lambda\) and \(k\), consistently surpassing the baseline. For simplicity, we employ a moderate range of hyperparameters, consistently setting \(\lambda = 0.5\) and \(k = 1\) throughout the experiments. Further tuning these parameters on a hold-out set may further enhance the performance of conformal prediction.

\paragraph{\texttt{SACP} is robust to the size of calibration set} 
As shown in Eq. \eqref{eq:calculate_tau}, the calibration set is used to determine the threshold \(\tau\) corresponding to the coverage rate of \(1-\alpha\). A precise estimate of $\tau$ is essential. In this work, all experiments are conducted with a fixed ratio for the calibration set size, specifically \( |\mathcal{D}_{\text{cal}}| \approx 0.5 \times |\mathcal{D}_{\text{cal}} \cup \mathcal{D}_{\text{test}}| \). In this section, we investigate whether \texttt{SACP} is sensitive to the size of the calibration set \( |\mathcal{D}_{\text{cal}}| \). Specifically, we exam whether  Coverage and Size  remain satisfactory when varying the ratio of the calibration set size. The experiment is conducted on all three datasets using SSTN, with APS as the base score function and the error rate \(\alpha\) set to $0.05$. The ratio of calibration samples is defined as \(\gamma = (|\mathcal{D}_{\text{cal}}| / |\mathcal{D}_{\text{cal}} \cup \mathcal{D}_{\text{test}}|)\). \(\gamma\) ranges from 0.1 to 0.6 in increments of 0.05. The experimental results are presented in Table \ref{tab:change}. Coverage consistently maintains \(1 - \alpha\) as the calibration set size decreases relative to \(|\mathcal{D}_{\text{cal}} \cup \mathcal{D}_{\text{test}}|\), even for very small values of \(\gamma\). Additionally, the size of the prediction sets remains stable, only marginally influenced by changes in the calibration set size, and fluctuates within a narrow range. These findings collectively demonstrate that \texttt{SACP} exhibits significant robustness to variations in the calibration set size.

\subsection{Discussion}

\paragraph{\texttt{SACP} exhibits adaptiveness}Adaptiveness is a crucial attribute of conformal prediction, ensuring that prediction sets accurately reflect the uncertainty associated with each instance. Specifically, prediction sets should be larger for more challenging instances and smaller for easier ones. This analysis evaluates whether \texttt{SACP} maintains this characteristic. We use the true label rank \(o(y, \hat{\pi}_\theta(\boldsymbol{x}))\) to represent the difficulty of the instance \(\boldsymbol{x}\), where a higher true label rank indicates greater classification difficulty. Instances are categorized into five groups based on their true label rank: \(1\), \(2\), \(3\), \(4\), and \(\{5, \ldots, |\mathcal{Y}|\}\). We then compute the average size of prediction sets for each group. This experiment is conducted on the SA and PU datasets using SSTN, with APS as the base score function and the error rate \(\alpha\) set to \(0.05\).

Figure \ref{fig:sizevsdiff} demonstrates that \texttt{SACP} exhibits adaptiveness across various datasets. In both cases, with or without \texttt{SACP}, APS consistently shows that prediction sets are smaller for easier instances compared to more challenging ones. We conclude that \texttt{SACP} effectively optimizes the average prediction set size for each difficulty group while preserving its adaptive characteristics. These results collectively underscore the significance of \texttt{SACP} in HSI classification.

\paragraph{\texttt{SACP}\label{dis:compu} is  computationally efficient}In this section, we analyze the computational efficiency of \texttt{SACP}. The base non-conformity score functions APS, RAPS, and SAPS involve only sorting and summation operations, resulting in a time complexity of \(\mathcal{O}(|\mathcal{Y}| \log |\mathcal{Y}| N)\), where \(N\) is the number of instances and \(|\mathcal{Y}|\) represents the total number of labels. The dominant computational overhead arises from the sorting operation. In comparison, \texttt{SACP} introduces an additional aggregation step, which adds a time complexity of \(\mathcal{O}(kN)\), where \(k\) is the number of iterations. Consequently, the overall time complexity of \texttt{SACP} is \(\mathcal{O}(|\mathcal{Y}| \log |\mathcal{Y}| N + kN)\). For simplicity, we approximate \((k + |\mathcal{Y}| \log |\mathcal{Y}|)\) as a constant \(C\), as these terms are negligible compared to \(N\). Therefore, the complexity can be expressed as \(\mathcal{O}(CN)\), which is comparable in magnitude to that of standard conformal prediction. This analysis demonstrates that \texttt{SACP} exhibits computational efficiency.

\section{Conclusion}
In this paper, we first present a comprehensive theoretical proof establishing the validity of conformal prediction for HSI classification. We then introduce a conformal procedure that integrates conformal prediction sets into HSI classifiers, ensuring the inclusion of true labels with a user-specified probability. Building on this foundation, we propose \texttt{SACP}, a conformal prediction framework specifically designed for HSI data. \texttt{SACP} refines non-conformity scores by leveraging the intrinsic spatial information present in HSIs, effectively enhancing the efficiency of conformal prediction sets. To the best of our knowledge, this is the first study to explore conformal prediction in the context of HSI data. Extensive experiments  confirm the validity of conformal prediction for HSI classification. Moreover, \texttt{SACP} outperforms standard conformal prediction in HSI classification by reducing the average prediction set size while maintaining satisfactory marginal and conditional coverage rates. Furthermore, \texttt{SACP} is practical and user-friendly, exhibiting robustness to hyperparameter variations and incurring minimal computational costs.

\noindent\textbf{Limitations.} The number of pixels within different classes in HSIs can sometimes be imbalanced, leading to unusually large prediction sets and potential undercoverage for classes with fewer pixels. This issue has not been addressed in our work and represents a promising avenue for future research.

\appendices
\section{Proof of Theorem \ref{theo:exchange}}
\label{appendix:th1}
\begin{proof}
Here, we emphasize that the non-conformity score \(S(B, y)\) is dependent on the sample partitioning, i.e., 
\begin{equation}
    S(B, y) \equiv S(B, y; \mathcal{D}_{\text{train}}, \mathcal{D}_{\text{val}}, \mathcal{D}_{\text{cal}}\cup \mathcal{D}_{\text{test}}).
\end{equation}
By Assumption \ref{ass:exchange}, the non-conformity score \( S(B, y) \) remains invariant under any permutation \(\pi\) of the union \( \mathcal{D}_{\text{cal}} \cup \mathcal{D}_{\text{test}} \). Therefore, we have 
\begin{equation}
\begin{aligned}
        &S(B, y; \mathcal{D}_{\text{train}}, \mathcal{D}_{\text{val}}, \mathcal{D}_{\text{cal}}\cup \mathcal{D}_{\text{test}}) \\= &S(B, y; \mathcal{D}_{\text{train}}, \mathcal{D}_{\text{val}}, (\mathcal{D}_{\text{cal}} \cup \mathcal{D}_{\text{test}})^{\pi}),
\end{aligned}
\end{equation}
where \((\mathcal{D}_{\text{cal}} \cup \mathcal{D}_{\text{test}})^{\pi}\) denotes permuted samples according to \(\pi\), and \(\pi\) is permutation of \(|\mathcal{D}_{\text{cal}} \cup \mathcal{D}_{\text{test}}|\). 
Hence, the score \( S(B, y) \) remains invariant with respect to the selection of any subset of \( \mathcal{D}_{\text{cal}} \cup \mathcal{D}_{\text{test}} \) as the calibration set \( \mathcal{D}_{\text{cal}} \). Let \([t_1, t_2, \ldots, t_{n+m}]\) be the unordered scores from \(\{S(B_i, y_i)\}_{i\in \mathcal{D}_{\text{cal}} \cup \mathcal{D}_{\text{test}}}\), where \([\cdot]\) denotes unordered sets. We have \(\{s_i\}_{i=1}^n\) is a subset of size \(n\) of \([t_1, t_2, \ldots, t_{n+m}]\). Under random splitting of \(\mathcal{D}_{\text{cal}} \cup \mathcal{D}_{\text{test}}\), any permutation \(\pi\) happens with the same probability, i.e., 
\begin{equation}
    \mathbb{P}\left(\{s_i\}_{i=1}^n = \{t_1, t_2, \ldots, t_n\} \mid \{s_i\}_{i=1}^{n+m}\right) = \frac{1}{\binom{n+m}{n}}.
\end{equation}
As a result, any score of test sample in  \(\{s_{n+j}\}_{j=1}^m\) is exchangeable with \(\{s_i\}_{i=1}^n\). 
\end{proof}

\section{Proof of Proposition \ref{pro:score}}
\label{appendix:pro1}
\begin{proof}We introduce the superscript \(\alpha\) to \(\tau\) as follows:
\begin{equation}
\label{eq:taual}
    \tau^\alpha = \inf \left\{ s \mid \frac{\left|\{i \mid s_i \leq s\}\right|}{N_2} \geq \frac{\lceil (N_2 + 1)(1 - \alpha) \rceil}{N_2} \right\}.
\end{equation}
Denote the cuboids in $\mathcal{D}_{\text{test}}$ as $\mathcal{B}_{\text{test}}$. Then, for each cuboid $B$ in $\mathcal{B}_{\text{test}}$, we have
\begin{equation}
\label{eq:begin}
\begin{aligned}
        \int_0^1 |\mathcal{C}_{1-\alpha}(B)|\mathrm{d}\alpha &= \int_0^1|\{y \in \mathcal{Y} \mid S(B, y) \leq \tau^\alpha\}|~~ \mathrm{d}\alpha\\&= \int_{0}^{1}\sum_{y\in \mathcal{Y}}\mathds{1}_{\{S(B, y))\le \tau^{\alpha}\}} ~~ \mathrm{d}\alpha.
\end{aligned}
\end{equation}
 Moreover, the event $\{S(B, y) \leq \tau^{\alpha}\}$ in Eq. \eqref{eq:begin} is equivalent to the event that at least $\alpha$ percent of the sample in $\mathcal{D}_{\text{cal}}$ have scores larger than that of $(B, y)$. Thus, we have
\begin{equation}
\begin{aligned}
      &\int_{0}^{1}\sum_{y\in \mathcal{Y}}\mathds{1}_{\{S(B, y))\le \tau^{\alpha}\}} ~~ \mathrm{d}\alpha \\ = & \int_{0}^{1}\sum_{y\in \mathcal{Y}}\mathds{1}_{\{   \sum_{(\tilde{B}, \tilde{y})\in \mathcal{D}_{\text{cal}}}\mathds{1}_{\{S(\tilde{B}, \tilde{y}) \ge S(B, y)\}}  \ge \lfloor (1 + |\mathcal{D}_{\text{cal}}|)\alpha\rfloor \}} ~~ \mathrm{d}\alpha \\ = &  \int_{0}^{1}\sum_{y\in \mathcal{Y}}\mathds{1}_{\{\frac{1+    \sum_{(\tilde{B}, \tilde{y})\in \mathcal{D}_{\text{cal}}}\mathds{1}_{\{S(\tilde{B}, \tilde{y}) \ge S(B, y)\}} }{1 + |\mathcal{D}_{\text{cal}}|} > \alpha\}} ~~ \mathrm{d}\alpha.
\end{aligned}
\end{equation}
It is obvious that 
\begin{equation}
    0 < \frac{1+    \sum_{(\tilde{B}, \tilde{y})\in \mathcal{D}_{\text{cal}}}\mathds{1}_{\{S(\tilde{B}, \tilde{y}) \ge S(B, y)\}} }{1 + |\mathcal{D}_{\text{cal}}|}\le 1.
\end{equation}
Thus, we have 
\begin{equation}
\begin{aligned}
        &\int_{0}^{1}\sum_{y\in \mathcal{Y}}\mathds{1}_{\{\frac{1+    \sum_{(\tilde{B}, \tilde{y})\in \mathcal{D}_{\text{cal}}}\mathds{1}_{\{S(\tilde{B}, \tilde{y}) \ge S(B, y)\}} }{1 + |\mathcal{D}_{\text{cal}}|} > \alpha\}} ~~ \mathrm{d}\alpha \\= &\sum_{y\in \mathcal{Y}} \int_{0}^{1}\mathds{1}_{\{\frac{1+    \sum_{(\tilde{B}, \tilde{y})\in \mathcal{D}_{\text{cal}}}\mathds{1}_{\{S(\tilde{B}, \tilde{y}) \ge S(B, y)\}} }{1 + |\mathcal{D}_{\text{cal}}|} > \alpha\}} ~~ \mathrm{d}\alpha \\ = &\sum_{y\in \mathcal{Y}}\frac{1+    \sum_{(\tilde{B}, \tilde{y})\in \mathcal{D}_{\text{cal}}}\mathds{1}_{\{S(\tilde{B}, \tilde{y}) \ge S(B, y)\}} }{1 + |\mathcal{D}_{\text{cal}}|}
\end{aligned}
\end{equation}
 For $\mathcal{B}_{\text{test}}$, we have
\begin{equation}
\begin{aligned}
        &\sum_{B \in \mathcal{B}_{\text{test}}} \int_0^1 |\mathcal{C}_{1-\alpha}(B)| \, \mathrm{d}\alpha \\=& \sum_{B \in \mathcal{B}_{\text{test}}}\sum_{y\in \mathcal{Y}}\frac{1+    \sum_{(\tilde{B}, \tilde{y})\in \mathcal{D}_{\text{cal}}}\mathds{1}_{\{S(\tilde{B}, \tilde{y}) \ge S(B, y)\}} }{1 + |\mathcal{D}_{\text{cal}}|}\\=&\sum_{(B,y)\in \hat{\mathcal{D}}_{\text{test}}}\frac{1+    \sum_{(\tilde{B}, \tilde{y})\in \mathcal{D}_{\text{cal}}}\mathds{1}_{\{S(\tilde{B}, \tilde{y}) \ge S(B, y)\}} }{1 + |\mathcal{D}_{\text{cal}}|}\\=&\sum_{(B,y)\in \hat{\mathcal{D}}_{\text{test}}}\frac{1+    \sum_{(\tilde{B}, \tilde{y})\in \mathcal{D}_{\text{cal}}}\mathds{1}_{\{S(\tilde{B}, \tilde{y}) \ge S(B, y)\}} }{1 + n}\\=&\frac{n'|\mathcal{Y}|}{1 + n} + \frac{1}{1 + n }\sum_{i=1}^{n} \sum_{k=1}^{n'|\mathcal{Y}|} \mathds{1}_{\{s_i > t_k\}}.
\end{aligned}
\end{equation}
 In addition, we have
\begin{equation}
    {R}(\{s_i\}_{i=1}^{n},\{t_k\}_{k=1}^{n'|\mathcal{Y}|}) = \frac{1}{n \times n'|\mathcal{Y}|} \sum_{i=1}^{n} \sum_{k=1}^{n'|\mathcal{Y}|} \mathds{1}_{\{s_i > t_k\}},
\end{equation}
Finally, we have 
\begin{equation}
\label{final}
\begin{aligned}
        &\sum_{(B,y) \in \mathcal{D}_{\text{test}}} \int_0^1 |\mathcal{C}_{1-\alpha}(B)| \, \mathrm{d}\alpha \\= &\frac{n \times n'|\mathcal{Y}|}{1+n}{R}(\{s_i\}_{i=1}^{n},\{t_k\}_{k=1}^{n'|\mathcal{Y}|}) + \frac{n'|\mathcal{Y}|}{1 + n}, 
\end{aligned}
\end{equation}
where $\frac{n \times n'|\mathcal{Y}|}{1+n}$ and $\frac{n'|\mathcal{Y}|}{1 + n}$ are positive constants. Based on Eq. \eqref{final}, we obviously have

\small
\begin{equation}
\begin{aligned}
     {R}(\{s_i\}_{i=1}^{n},\{t_k\}_{k=1}^{n'|\mathcal{Y}|}) &> {R}(\{\hat{{s}}_i\}_{i=1}^{n},\{\hat{t}_k\}_{k=1}^{n'|\mathcal{Y}|}) \Leftrightarrow \\
     \sum_{(B,y) \in \mathcal{D}_{\text{test}}} \int_0^1 |\mathcal{C}_{1-\alpha}(B)| \, \mathrm{d}\alpha &> \sum_{(B,y) \in \mathcal{D}_{\text{test}}} \int_0^1 |\hat{\mathcal{C}}_{1-\alpha}(B)| \, \mathrm{d}\alpha.
\end{aligned}
\end{equation}
\normalsize
Then, we complete the proof.
\end{proof}

\section{Proof of Proposition \ref{pro:sacpcon}}
\label{appendix:pro2}
\begin{proof} We will employ mathematical induction to demonstrate that the statement is valid for all \( k \).

\paragraph{Base Case (\(k = 0\))}
For \(k = 0\), the operator \(\mathcal{V}_k\) reduces to \(S\). By our Theorem \ref{theo:exchange}, we have $\hat{s}_{i}, (i =1, \cdots, n)$ and $S(B_{n+1}, y_{n+1})$ are exchangeable. Thus, 
\begin{equation}
\begin{aligned}
        &\mathbb{P}\left(y_{n+1} \notin \hat{\mathcal{C}}_{1-\alpha}(B_{n+1}; \hat{\tau})\right) \\= & \mathbb{P}\left(S(B_{n+1}, y_{n+1}) > \hat{\tau}\right) \\
        = & \frac{1}{n+1} \sum_{i=1}^{n+1} \mathbb{P}\left(S(B_{i}, y_{i}) > \hat{\tau}\right) \\
        = & \frac{1}{n+1} \sum_{i=1}^{n+1} \mathbb{E} \mathds{1}_{\{ i: S(B_{i}, y_{i}) > \hat{\tau} \} }\\ 
        = & \frac{1}{n+1} \mathbb{E} |\{ i: S(B_{i}, y_{i}) > \hat{\tau} \}| \\ 
        \le& \frac{1}{n+1} (n+1)\alpha\\ 
        = & \alpha.
\end{aligned}
\end{equation}
Therefore, we have
\begin{equation}
    \mathbb{P}\left(y_{n+1} \in \hat{\mathcal{C}}_{1-\alpha}(B_{n+1}; \hat{\tau})\right) \ge 1-\alpha
\end{equation}

\paragraph{Inductive Hypothesis}
Assume the proposition holds for \(k = m\). Specifically, assume:
\begin{equation}
\mathbb{P}\left(y_{n+1} \in \hat{\mathcal{C}}_{1-\alpha}(B_{n+1}; \hat{\tau})\right) \geq 1 - \alpha,
\end{equation}
where 
\begin{equation}
\hat{\mathcal{C}}_{1-\alpha}(B_{n+1}; \hat{\tau}) := \{y \in \mathcal{Y} \mid \mathcal{V}_m(B_{n+1}, y) \leq \hat{\tau}\},
\end{equation}
and $\hat{s}_{i}^{(m)}, i =1, \dots, n+1$ are exchangeable. 

\paragraph{Inductive Step}
To prove the proposition for \(k = m + 1\), we need to show:
\begin{equation}
\mathbb{P}\left(y_{n+1} \in \hat{\mathcal{C}}_{1-\alpha}(B_{n+1}; \hat{\tau})\right) \geq 1 - \alpha,
\end{equation}
where 
\begin{equation}
\hat{\mathcal{C}}_{1-\alpha}(B_{n+1}; \hat{\tau}) := \{y \in \mathcal{Y} \mid \mathcal{V}_{m+1}(B_{n+1}, y) \leq \hat{\tau}\},
\end{equation}
and $\hat{s}_{i}^{(m+1)}, i =1, \dots, n+1$ are exchangeable. 

The aggregation operator for \(k = m + 1\) is:
\begin{equation}
\label{eq:agg}
\mathcal{V}_{m+1}(B_i, y) = (1 - \lambda) \mathcal{V}_m(B_i, y) + \frac{\lambda}{|\mathcal{N}_i|} \sum_{B_j \in \mathcal{N}_i} \mathcal{V}_m(B_j, y),
\end{equation}where \(\mathcal{N}_i\) denotes the set of neighboring cuboids around the central cuboid \(B_i\) included in \(\mathcal{B}\). Here, \(\mathcal{B}\) represents the set of all exchangeable cuboids of an HSI to be aggregated. 
Let \(g(\hat{s}_{1}^{(m)}, \dots, \hat{s}_{n+1}^{(m)}) = (\hat{s}_{1}^{(m+1)}, \dots, \hat{s}_{n+1}^{(m+1)})^\top\). Since the operator \(g(\cdot)\) preserves exchangeability by Theorem 3 in \cite{kuchibhotla2021exchangeability}, we have the aggregated scores\(\hat{s}_{1}^{(m+1)}, \dots, \hat{s}_{n+1}^{(m+1)}\) are still exchangeable. Correspondingly, the coverage guarantee is maintained for \(k = m + 1\). By Mathematical Induction, the proposition holds for all non-negative integers \(k\). 
\end{proof}

\bibliographystyle{IEEEtran}
\bibliography{ref}

\end{document}